# Exploiting Structure in Weighted Model Counting Approaches to Probabilistic Inference

**Wei Li** WEI.LI@AUTODESK.COM
*Autodesk Canada*
*Toronto, Ontario M5A 1J7 Canada*
**Pascal Poupart** PPOUPART@CS.UWATERLOO.CA
**Peter van Beek** VANBEEK@CS.UWATERLOO.CA
*Cheriton School of Computer Science*
*University of Waterloo*
*Waterloo, Ontario N2L 3G1 Canada*

## Abstract

Previous studies have demonstrated that encoding a Bayesian network into a SAT formula and then performing weighted model counting using a backtracking search algorithm can be an effective method for exact inference. In this paper, we present techniques for improving this approach for Bayesian networks with noisy-OR and noisy-MAX relations—two relations that are widely used in practice as they can dramatically reduce the number of probabilities one needs to specify. In particular, we present two SAT encodings for noisy-OR and two encodings for noisy-MAX that exploit the structure or semantics of the relations to improve both time and space efficiency, and we prove the correctness of the encodings. We experimentally evaluated our techniques on large-scale real and randomly generated Bayesian networks. On these benchmarks, our techniques gave speedups of up to two orders of magnitude over the best previous approaches for networks with noisy-OR/MAX relations and scaled up to larger networks. As well, our techniques extend the weighted model counting approach for exact inference to networks that were previously intractable for the approach.

## 1. Introduction

Bayesian networks are a fundamental building block of many AI applications. A Bayesian network consists of a directed acyclic graph where the nodes represent the random variables and each node is labeled with a conditional probability table (CPT) that represents the strengths of the influences of the parent nodes on the child node (Pearl, 1988). In general, assuming random variables with domain size $d$, the CPT of a child node with $n$ parents requires one to specify $d^{n+1}$ probabilities. This presents a practical difficulty and has led to the introduction of patterns for CPTs that require one to specify many fewer parameters (e.g., Good, 1961; Pearl, 1988; Díez & Druzdzel, 2006).

Perhaps the most widely used patterns in practice are the noisy-OR relation and its generalization, the noisy-MAX relation (Good, 1961; Pearl, 1988). These relations assume a form of causal independence and allow one to specify a CPT with just $n$ parameters in the case of the noisy-OR and $(d-1)^2 n$ parameters in the case of the noisy-MAX, where $n$ is the number of parents of the node and $d$ is the size of the domains of the random variables. The noisy-OR/MAX relations have been successfully applied in the knowledge engineering of large real-world Bayesian networks, such as the Quick Medical Reference-Decision Theoretic (QMR-DT) project (Miller, Masarie, & Myers, 1986) and the Computer-based Patient Case Simulation system (Parker & Miller, 1987). As well, Zagorecki and Druzdzel (1992) show that in three real-world Bayesian networks, noisy-OR/MAX relations were a good fit for up to 50% of the CPTs in these networks and that converting some CPTs to noisy-OR/MAX relations gave good approximations when answering probabilistic queries. This is surprising, as the CPTs in these networks were not specified using the noisy-OR/MAX assumptions and were specified as full CPTs. Their results provide additional evidence for the usefulness of noisy-OR/MAX relations.

We consider here the problem of exact inference in Bayesian networks that contain noisy-OR/MAX relations. One method for solving such networks is to replace each noisy-OR/MAX by its full CPT representation and then use any of the well-known algorithms for answering probabilistic queries such as variable elimination or tree clustering/jointree. However, in general—and in particular, for the networks that we use in our experimental evaluation—this method is impractical. A more fruitful approach for solving such networks is to take advantage of the semantics of the noisy-OR/MAX relations to improve both time and space efficiency (e.g., Heckerman, 1989; Olesen, Kjaerulff, Jensen, Jensen, Falck, Andreassen, & Andersen, 1989; D'Ambrosio, 1994; Heckerman & Breese, 1996; Zhang & Poole, 1996; Takikawa & D'Ambrosio, 1999; Díez & Galán, 2003; Chavira, Allen, & Darwiche, 2005).

Previous studies have demonstrated that encoding a Bayesian network into a SAT formula and then performing weighted model counting using a DPLL-based algorithm can be an effective method for exact inference, where DPLL is a backtracking algorithm specialized for SAT that includes unit propagation, conflict recording, backjumping, and component caching (Sang, Beame, & Kautz, 2005a). In this paper, we present techniques for improving this weighted model counting approach for Bayesian networks with noisy-OR and noisy-MAX relations. In particular, we present two CNF encodings for noisy-OR and two CNF encodings for noisy-MAX that exploit their semantics to improve both the time and space efficiency of probabilistic inference. In our encodings, we pay particular attention to reducing the treewidth of the CNF formula. We also explore alternative search ordering heuristics for the DPLL-based backtracking algorithm.

We experimentally evaluated our encodings on large-scale real and randomly generated Bayesian networks using the Cachet weighted model counting solver (Sang, Bacchus, Beame, Kautz, & Pitassi, 2004). While our experimental results must be interpreted with some care as we are comparing not only our encodings but also implementations of systems with conflicting design goals, on these benchmarks our techniques gave speedups of up to three orders of magnitude over the best previous approaches for networks with noisy-OR and noisy-MAX. As well, on these benchmarks there were many networks that could not be solved at all by previous approaches within resource limits, but could be solved quite quickly by Cachet using our encodings. Thus, our noisy-OR and noisy-MAX encodings extend the model counting approach for exact inference to networks that were previously intractable for the approach.

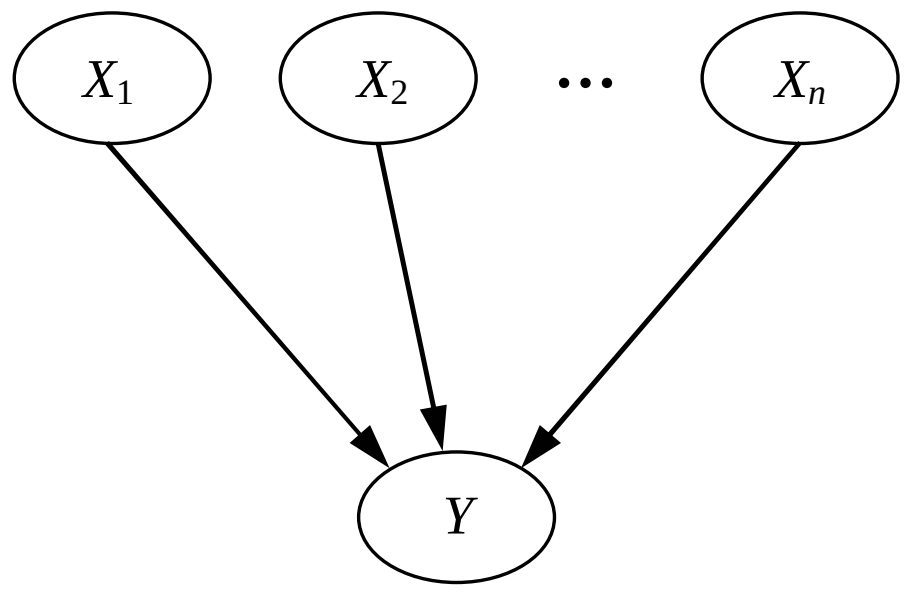

Figure 1: General causal structure for a Bayesian network with a noisy-OR/MAX relation, where causes $X_1, \ldots, X_n$ lead to effect $Y$ and there is a noisy-OR/MAX relation at node $Y$.

## 2. Background

In this section, we review noisy-OR/MAX relations and the needed background on weighted model counting approaches to exact inference in Bayesian networks (for more on these topics see, for example, Koller & Friedman, 2009; Darwiche, 2009; Chavira & Darwiche, 2008).

### 2.1 Patterns for CPTs: Noisy-OR and Noisy-MAX

With the noisy-OR relation one assumes that there are different causes $X_1, \ldots, X_n$ leading to an effect $Y$ (see Figure 1), where all random variables are assumed to have Boolean-valued domains. Each cause $X_i$ is either present or absent, and each $X_i$ in isolation is likely to cause $Y$ and the likelihood is not diminished if more than one cause is present. Further, one assumes that all possible causes are given and when all causes are absent, the effect is absent. Finally, one assumes that the mechanism or reason that inhibits a $X_i$ from causing $Y$ is independent of the mechanism or reason that inhibits a $X_j$, $j \neq i$, from causing $Y$.

A noisy-OR relation specifies a CPT using $n$ parameters, $q_1, \ldots, q_n$, one for each parent, where $q_i$ is the probability that $Y$ is false given that $X_i$ is true and all of the other parents are false,

$$P(Y = 0 \mid X_i = 1, X_j = 0_{[\forall j, j \neq i]}) = q_i. \tag{1}$$

From these parameters, the full CPT representation of size $2^{n+1}$ can be generated using,

$$P(Y = 0 \mid X_1, \ldots, X_n) = \prod_{i \in T_x} q_i \tag{2}$$

and

$$P(Y = 1 \mid X_1, \ldots, X_n) = 1 - \prod_{i \in T_x} q_i \tag{3}$$

where $T_x = \{i \mid X_i = 1\}$ and $P(Y = 0 \mid X_1, \ldots, X_n) = 1$ if $T_x$ is empty. The last condition (when $T_x$ is empty) corresponds to the assumptions that all possible causes are given and that when all causes are absent, the effect is absent; i.e., $P(Y = 0 \mid X_1 = 0, \ldots, X_n = 0) = 1$. These assumptions are not as restrictive as may first appear. One can always introduce an additional random variable $X_0$ that is a parent of $Y$ but itself has no parents. The variable $X_0$ represents all of the other reasons that could cause $Y$ to occur. The node $X_0$ and the prior probability $P(X_0)$ are referred to as a *leak node* and the *leak probability*, respectively. In what follows, we continue to refer to all possible causes as $X_1, \ldots, X_n$ where it is understood that one of these causes could be a leak node.

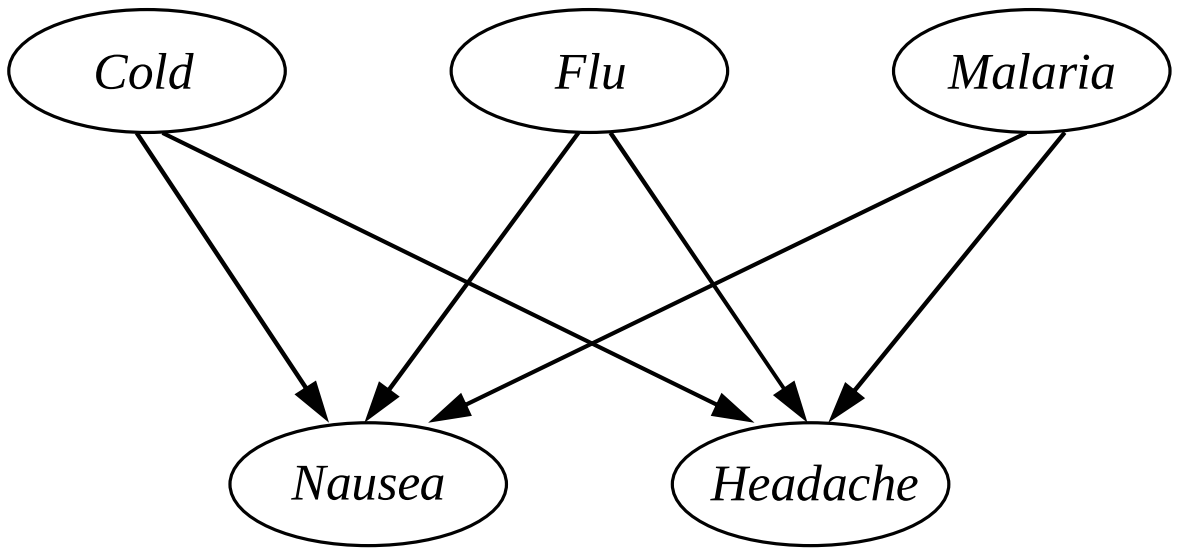

Figure 2: Example of a causal Bayesian network with causes (diseases) *Cold*, *Flu*, and *Malaria* and effects (symptoms) *Nausea* and *Headache*.

**Example 1.** *Consider the Bayesian network shown in Figure 2. Suppose that the random variables are Boolean representing the presence or the absence of the disease or the symptom, that there is a noisy-OR at node Nausea and at node Headache, and that the parameters for the noisy-ORs are as given in Table 1. The full CPT for the node Nausea is given by,*

| $C$ | $F$ | $M$ | $P(Nausea = 0 \mid C, F, M)$ | $P(Nausea = 1 \mid C, F, M)$ |
|---|---|---|---|---|
| 0 | 0 | 0 | 1.00 | 0.00 |
| 0 | 0 | 1 | 0.40 | 0.60 |
| 0 | 1 | 0 | 0.50 | 0.50 |
| 0 | 1 | 1 | $0.20 = 0.5 \times 0.4$ | 0.80 |
| 1 | 0 | 0 | 0.60 | 0.40 |
| 1 | 0 | 1 | $0.24 = 0.6 \times 0.4$ | 0.76 |
| 1 | 1 | 0 | $0.30 = 0.6 \times 0.5$ | 0.70 |
| 1 | 1 | 1 | $0.12 = 0.6 \times 0.5 \times 0.4$ | 0.88 |

*where $C$, $F$, and $M$ are short for Cold, Flu, and Malaria, respectively.*

An alternative way to view a noisy-OR relation is as a decomposed probabilistic model. In the decomposed model shown in Figure 3, one only has to specify a small conditional probability table at each node $Y_i$ given by $P(Y_i \mid X_i)$, instead of an exponentially large CPT given by $P(Y \mid X_1, \ldots, X_n)$. In the decomposed model, $P(Y_i = 0 \mid X_i = 0) = 1$, $P(Y_i = 0 \mid X_i = 1) = q_i$, and the CPT at node $Y$ is now deterministic and is given by the OR logical relation. The OR operator can be converted into a full CPT as follows,

$$P(Y \mid Y_1, \ldots, Y_n) = \begin{cases} 1, & \text{if } Y = Y_1 \vee \cdots \vee Y_n, \\ 0, & \text{otherwise.} \end{cases}$$

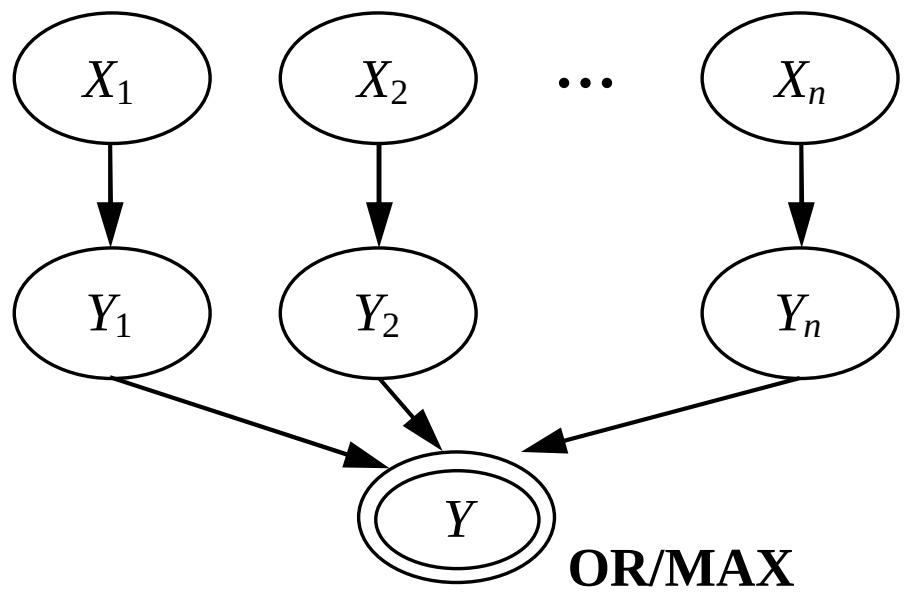

Figure 3: Decomposed form for a Bayesian network with a noisy-OR/MAX relation, where causes $X_1, \ldots, X_n$ lead to effect $Y$ and there is a noisy-OR/MAX relation at node $Y$. The node with a double border is a deterministic node with the designated logical relationship (OR) or arithmetic relationship (MAX).

Table 1: Parameters for the noisy-ORs at node *Nausea* and at node *Headache* for the Bayesian network shown in Figure 2, assuming all of the random variables are Boolean.

$$\begin{aligned}
P(\mathit{Nausea} = 0 \mid \mathit{Cold} = 1, \mathit{Flu} = 0, \mathit{Malaria} = 0) &= 0.6 \\
P(\mathit{Nausea} = 0 \mid \mathit{Cold} = 0, \mathit{Flu} = 1, \mathit{Malaria} = 0) &= 0.5 \\
P(\mathit{Nausea} = 0 \mid \mathit{Cold} = 0, \mathit{Flu} = 0, \mathit{Malaria} = 1) &= 0.4 \\
P(\mathit{Headache} = 0 \mid \mathit{Cold} = 1, \mathit{Flu} = 0, \mathit{Malaria} = 0) &= 0.3 \\
P(\mathit{Headache} = 0 \mid \mathit{Cold} = 0, \mathit{Flu} = 1, \mathit{Malaria} = 0) &= 0.2 \\
P(\mathit{Headache} = 0 \mid \mathit{Cold} = 0, \mathit{Flu} = 0, \mathit{Malaria} = 1) &= 0.1
\end{aligned}$$

The probability distribution of an effect variable $Y$ is given by,

$$P(Y \mid X_1, \ldots, X_n) = \sum_{Y = Y_1 \vee \cdots \vee Y_n} \left( \prod_{i=1}^{n} P(Y_i \mid X_i) \right),$$

where the sum is over all configurations or possible values for $Y_1, \ldots, Y_n$ such that the OR of these Boolean values is equal to the value for $Y$. Similarly, in Pearl's (1988) decomposed model, one only has to specify $n$ probabilities to fully specify the model (see Figure 4); i.e., one specifies the prior probabilities $P(I_i)$, $1 \leq i \leq n$. In this model, causes always lead to effects unless they are prevented or inhibited from doing so. The random variables $I_i$ model this prevention or inhibition.

These two decomposed probabilistic models (Figure 3, and Figure 4) can be shown to be equivalent in the sense that the conditional probability distribution $P(Y \mid X_1, \ldots, X_n)$ induced by both of these networks is just the original distribution for the network shown in Figure 1. It is important to note that both of these models would still have an exponentially large CPT associated with the effect node $Y$ if the deterministic OR node were to be replaced by its full CPT representation. In other words, these decomposed models address ease of modeling or representation issues, but do not address efficiency of reasoning issues.

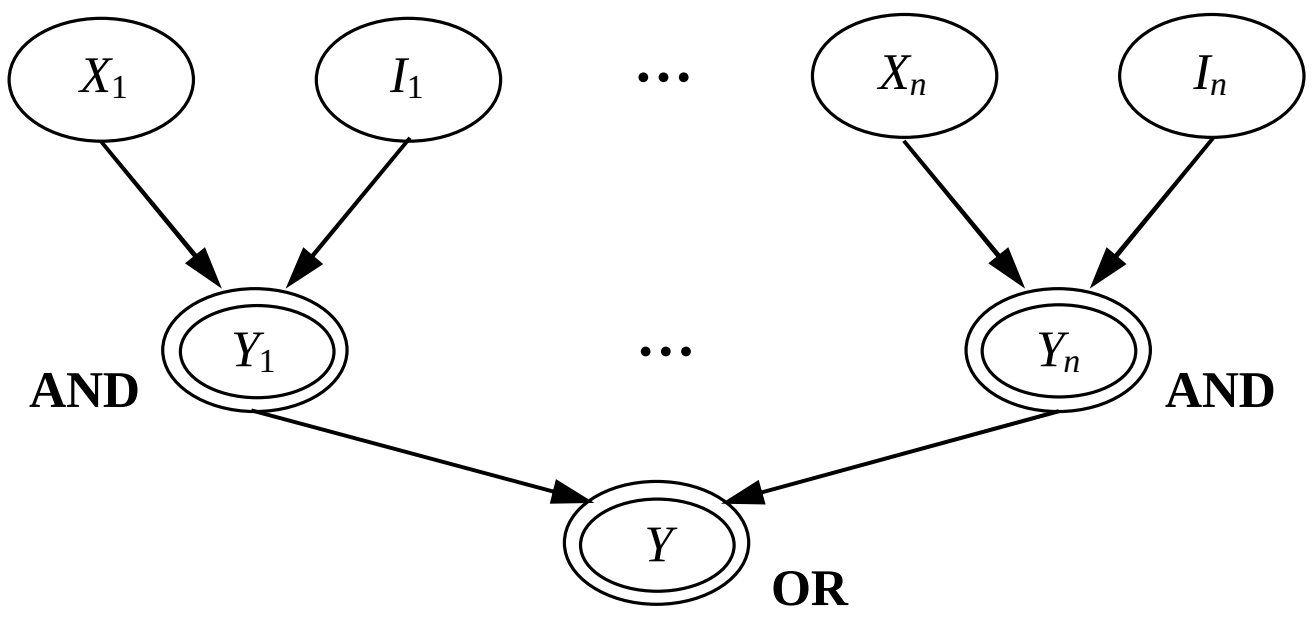

Figure 4: Pearl's (1988) decomposed form of the noisy-OR relation. Nodes with double borders are deterministic nodes with the designated logical relationship.

The noisy-MAX relation (see Pearl, 1988; Good, 1961; Henrion, 1987; Díez, 1993) is a generalization of the noisy-OR to non-Boolean domains. With the noisy-MAX relation, one again assumes that there are different causes $X_1, \ldots, X_n$ leading to an effect $Y$ (see Figure 1), but now the random variables may have multi-valued (non-Boolean) domains. The domains of the variables are assumed to be ordered and the values are referred to as the degree or the severity of the variable. Each domain has a distinguished lowest degree 0 representing the fact that a cause or effect is absent. As with the noisy-OR relation, one assumes that all possible causes are given and when all causes are absent, the effect is absent. Again, these assumptions are not as restrictive as first appears, as one can incorporate a leak node. As well, one assumes that the mechanism or reason that inhibits a $X_i$ from causing $Y$ is independent of the mechanism or reason that inhibits a $X_j$, $j \neq i$, from causing $Y$.

Let $d_X$ be the number of values in the domain of some random variable $X$. For simplicity of notation and without loss of generality, we assume that the domain of a variable $X$ is given by the set of integers $\{0, 1, \ldots, d_X - 1\}$. A noisy-MAX relation with causes $X_1, \ldots, X_n$ and effect $Y$ specifies a CPT using the parameters,

$$P(Y = y \mid X_i = x_i, X_j = 0_{[\forall j, j \neq i]}) = q_{i,y}^{x_i} \qquad \begin{array}{l} i = 1, \ldots, n, \\ y = 0, \ldots, d_Y - 1, \\ x_i = 1, \ldots, d_{X_i} - 1. \end{array} \tag{4}$$

If all of the domain sizes are equal to $d$, a total of $(d-1)^2 n$ non-redundant probabilities must be specified. From these parameters, the full CPT representation of size $d^{n+1}$ can be generated using,

$$P(Y \leq y \mid \boldsymbol{X}) = \prod_{\substack{i=1 \\ x_i \neq 0}}^{n} \sum_{y'=0}^{y} q_{i,y'}^{x_i} \tag{5}$$

and

$$P(Y = y \mid \boldsymbol{X}) = \begin{cases} P(Y \leq 0 \mid \boldsymbol{X}) & \text{if } y = 0, \\ P(Y \leq y \mid \boldsymbol{X}) - P(Y \leq y - 1 \mid \boldsymbol{X}) & \text{if } y > 0. \end{cases} \tag{6}$$

where $\boldsymbol{X}$ represents a certain configuration of the parents of $Y$, $\boldsymbol{X} = x_1, \ldots, x_n$, and $P(Y = 0 \mid X_1 = 0, \ldots, X_n = 0) = 1$; i.e., if all causes are absent, the effect is absent.

Table 2: Parameters for the noisy-MAX at node *Nausea* for the Bayesian network shown in Figure 2, assuming the diseases are Boolean random variables and the symptom *Nausea* has domain {absent = 0, mild = 1, severe = 2}.

$$
\begin{aligned}
P(\mathit{Nausea} = \text{absent} \mid \mathit{Cold} = 1, \mathit{Flu} = 0, \mathit{Malaria} = 0) &= 0.7 \\
P(\mathit{Nausea} = \text{mild} \mid \mathit{Cold} = 1, \mathit{Flu} = 0, \mathit{Malaria} = 0) &= 0.2 \\
P(\mathit{Nausea} = \text{severe} \mid \mathit{Cold} = 1, \mathit{Flu} = 0, \mathit{Malaria} = 0) &= 0.1 \\
P(\mathit{Nausea} = \text{absent} \mid \mathit{Cold} = 0, \mathit{Flu} = 1, \mathit{Malaria} = 0) &= 0.5 \\
P(\mathit{Nausea} = \text{mild} \mid \mathit{Cold} = 0, \mathit{Flu} = 1, \mathit{Malaria} = 0) &= 0.2 \\
P(\mathit{Nausea} = \text{severe} \mid \mathit{Cold} = 0, \mathit{Flu} = 1, \mathit{Malaria} = 0) &= 0.3 \\
P(\mathit{Nausea} = \text{absent} \mid \mathit{Cold} = 0, \mathit{Flu} = 0, \mathit{Malaria} = 1) &= 0.1 \\
P(\mathit{Nausea} = \text{mild} \mid \mathit{Cold} = 0, \mathit{Flu} = 0, \mathit{Malaria} = 1) &= 0.4 \\
P(\mathit{Nausea} = \text{severe} \mid \mathit{Cold} = 0, \mathit{Flu} = 0, \mathit{Malaria} = 1) &= 0.5
\end{aligned}
$$

**Example 2.** *Consider once again the Bayesian network shown in Figure 2. Suppose that the diseases are Boolean random variables and the symptoms Nausea and Headache have domains {absent = 0, mild = 1, severe = 2}, there is a noisy-MAX at node Nausea and at node Headache, and the parameters for the noisy-MAX at node Nausea are as given in Table 2. The full CPT for the node Nausea is given by,*

| $C$ | $F$ | $M$ | $P(N = a \mid C, F, M)$ | $P(N = m \mid C, F, M)$ | $P(N = s \mid C, F, M)$ |
|---|---|---|---|---|---|
| 0 | 0 | 0 | 1.000 | 0.000 | 0.000 |
| 0 | 0 | 1 | 0.100 | 0.400 | 0.500 |
| 0 | 1 | 0 | 0.500 | 0.200 | 0.300 |
| 0 | 1 | 1 | $0.050 = 0.5 \times 0.1$ | 0.300 | 0.650 |
| 1 | 0 | 0 | 0.700 | 0.200 | 0.100 |
| 1 | 0 | 1 | $0.070 = 0.7 \times 0.1$ | 0.380 | 0.550 |
| 1 | 1 | 0 | $0.350 = 0.7 \times 0.5$ | 0.280 | 0.370 |
| 1 | 1 | 1 | $0.035 = 0.7 \times 0.5 \times 0.1$ | 0.280 | 0.685 |

*where $C$, $F$, $M$, and $N$ are short for the variables Cold, Flu, Malaria, and Nausea, respectively, and a, m, and s are short for the values absent, mild, and severe, respectively. As an example calculation,* $P(\mathit{Nausea} = \mathit{mild} \mid \mathit{Cold} = 0, \mathit{Flu} = 1, \mathit{Malaria} = 1) = ((0.5 + 0.2) \times (0.1 + 0.4)) - (0.05) = 0.3$ *As a second example,* $P(\mathit{Nausea} = \mathit{mild} \mid \mathit{Cold} = 1, \mathit{Flu} = 1, \mathit{Malaria} = 1) = ((0.7 + 0.2) \times (0.5 + 0.2) \times (0.1 + 0.4)) - (0.035) = 0.28$

As with the noisy-OR relation, an alternative view of a noisy-MAX relation is as a decomposed probabilistic model (see Figure 3). In the decomposed model, one only has to specify a small conditional probability table at each node $Y_i$ given by $P(Y_i \mid X_i)$, where $P(Y_i = 0 \mid X_i = 0) = 1$ and $P(Y_i = y \mid X_i = x) = q_{i,y}^{x}$. Each $Y_i$ models the effect of the cause $X_i$ on the effect $Y$ in isolation; i.e., the degree or the severity of the effect in the case where only the cause $X_i$ is not absent and all other causes are absent. The CPT at node $Y$ is now deterministic and is given by the MAX arithmetic relation. This corresponds to the assumption that the severity or the degree reached by the effect $Y$ is the maximum of the degrees produced by each cause if they were acting independently; i.e., the maximum of the $Y_i$'s. This assumption is only valid if the effects do not accumulate. The MAX operator can be converted into a full CPT as follows,

$$P(Y \mid Y_1, \ldots, Y_n) = \begin{cases} 1, & \text{if } Y = \max\{Y_1, \ldots, Y_n\}, \\ 0, & \text{otherwise.} \end{cases}$$

The probability distribution of an effect variable $Y$ is given by,

$$P(Y \mid X_1, \ldots, X_n) = \sum_{Y = \max\{Y_1, \ldots, Y_n\}} \left( \prod_{i=1}^{n} P(Y_i \mid X_i) \right),$$

where the sum is over all configurations or possible values for $Y_1, \ldots, Y_n$ such that the maximum of these values is equal to the value for $Y$. In both cases, however, making the CPTs explicit is often not possible in practice, as their size is exponential in the number of causes and the number of values in the domains of the random variables.

### 2.2 Weighted Model Counting for Probabilistic Inference

In what follows, we consider propositional formulas in conjunctive normal form (CNF). A *literal* is a Boolean variable (also called a *proposition*) or its negation and a *clause* is a disjunction of literals. A clause with one literal is called a *unit clause* and the literal in the unit clause is called a *unit literal*. A propositional formula $F$ is in *conjunctive normal form* if it is a conjunction of clauses.

**Example 3.** *For example, $(x \vee \neg y)$ is a clause, and the formula,*

$$F = (x \vee \neg y) \wedge (x \vee y \vee z) \wedge (y \vee \neg z \vee w) \wedge (\neg w \vee \neg z \vee v) \wedge (\neg v \vee u),$$

*is in CNF, where $u$, $v$, $w$, $x$, $y$, and $z$ are propositions.*

Given a propositional formula in conjunctive normal form, the problem of determining whether there exists a variable assignment that makes the formula evaluate to *true* is called the *Boolean satisfiability problem* or *SAT*. A variable assignment that makes a formula evaluate to *true* is also called a *model*. The problem of counting the number of models of a formula is called *model counting*.

Let $F$ denote a propositional formula. We use the value 0 interchangeably with the Boolean value *false* and the value 1 interchangeably with the Boolean value *true*. The notation $F|_{v=false}$ represents a new formula, called the *residual formula*, obtained by removing all clauses that contain the literal $\neg v$ (as these clauses evaluate to *true*) and deleting the literal $v$ from all clauses. Similarly, the notation $F|_{v=true}$ represents the residual formula obtained by removing all clauses that contain the literal $v$ and deleting the literal $\neg v$ from all clauses. Let $s$ be a set of instantiated variables in $F$. The residual formula $F|_s$ is obtained by cumulatively reducing $F$ by each of the variables in $s$.

**Example 4.** *Consider once again the propositional formula $F$ given in Example 3. Suppose $x$ is assigned* false. *The residual formula is given by,*

$$F|_{x=0} = (\neg y) \wedge (y \vee z) \wedge (y \vee \neg z \vee w) \wedge (\neg w \vee \neg z \vee v) \wedge (\neg v \vee u).$$

As is clear, a CNF formula is satisfied if and only if each of its clauses is satisfied and a clause is satisfied if and only if at least one of its literals is equivalent to 1. In a unit clause, there is no choice and the value of the literal is said to be *forced* or *implied*. The process of *unit propagation* assigns all unit literals to the value 1. As well, the formula is simplified by removing the variables of the unit literals from the remaining clauses and removing clauses that evaluate to *true* (i.e., the residual formula is obtained) and the process continues looking for new unit clauses and updating the formula until no unit clause remains.

**Example 5.** *Consider again the propositional formula $F|_{x=0}$ given in Example 4, where $x$ has been assigned* false. *The unit clause $(\neg y)$ forces $y$ to be assigned* false. *The residual formula is given by,*

$$F|_{x=0,y=0} = (z) \wedge (\neg z \vee w) \wedge (\neg w \vee \neg z \vee v) \wedge (\neg v \vee u).$$

*In turn, the unit clause $(z)$ forces $z$ to be assigned* true. *Similarly, the assignments $w = 1$, $v = 1$, and $u = 1$ are forced.*

There are natural polynomial-time reductions between the Bayesian inference problem and model counting problems (Bacchus, Dalmao, & Pitassi, 2003). In particular, exact inference in Bayesian networks can be reduced to the weighted model counting of CNFs (Darwiche, 2002; Littman, 1999; Sang et al., 2005a). Weighted model counting is a generalization of model counting.

A *weighted model counting problem* consists of a CNF formula $F$ and for each variable $v$ in $F$, a weight for each literal: $weight(v)$ and $weight(\neg v)$. Let $s$ be an assignment of a value to every variable in the formula $F$ that satisfies the formula; i.e., $s$ is a model of the formula. The weight of $s$ is the product of the weights of the literals in $s$. The solution of a weighted model counting problem is the sum of the weights of all satisfying assignments; i.e.,

$$weight(F) = \sum_{s} \prod_{l \in s} weight(l),$$

where the sum is over all possible models and the product is over the literals in that model.

Chavira and Darwiche (2002, 2008) proposed an encoding of a Bayesian network into weighted model counting of a propositional formula in conjunctive normal form. Chavira and Darwiche's encoding proceeds as follows. At each step, we illustrate the encoding using the Bayesian network shown in Figure 2. For simplicity, we assume the random variables are all Boolean and we omit the node *Headache*. To improve clarity, we refer to the random variables in the Bayesian network as "nodes" and reserve the word "variables" for the Boolean variables in the resulting propositional formula.

- For each value of each node in the Bayesian network, an *indicator variable* is created,

$$\begin{array}{lllllll} C & : & I_{C_0}, I_{C_1}, & \quad & M & : & I_{M_0}, I_{M_1}, \\ F & : & I_{F_0}, I_{F_1}, & & N & : & I_{N_0}, I_{N_1}. \end{array}$$

- For each node, *indicator clauses* are generated which ensure that in each model exactly one of the corresponding indicator variables for each node is true,

$$\begin{array}{lllllll} C & : & (I_{C_0} \vee I_{C_1}) \wedge (\neg I_{C_0} \vee \neg I_{C_1}), & \quad & M & : & (I_{M_0} \vee I_{M_1}) \wedge (\neg I_{M_0} \vee \neg I_{M_1}), \\ F & : & (I_{F_0} \vee I_{F_1}) \wedge (\neg I_{F_0} \vee \neg I_{F_1}), & & N & : & (I_{N_0} \vee I_{N_1}) \wedge (\neg I_{N_0} \vee \neg I_{N_1}). \end{array}$$

- For each conditional probability table (CPT) and for each parameter (probability) value in the CPT, a *parameter variable* is created,

$$\begin{array}{llllllll} C & : & P_{C_0}, & P_{C_1}, & \quad N & : & P_{N_0|C_0,F_0,M_0}, & P_{N_1|C_0,F_0,M_0}, \\ F & : & P_{F_0}, & P_{F_1}, & & & \ldots, & \ldots, \\ M & : & P_{M_0}, & P_{M_1}, & & & P_{N_0|C_1,F_1,M_1}, & P_{N_1|C_1,F_1,M_1}. \end{array}$$

- For each parameter variable, a *parameter clause* is generated. A parameter clause asserts that the conjunction of the corresponding indicator variables implies the parameter variable and vice-versa,

$$\begin{array}{llll} C & : & I_{C_0} \Leftrightarrow P_{C_0}, & I_{C_1} \Leftrightarrow P_{C_1}, \\ F & : & I_{F_0} \Leftrightarrow P_{F_0}, & I_{F_1} \Leftrightarrow P_{F_1}, \\ M & : & I_{M_0} \Leftrightarrow P_{M_0}, & I_{M_1} \Leftrightarrow P_{M_1}, \\ N & : & I_{C_0} \wedge I_{F_0} \wedge I_{M_0} \wedge I_{N_0} \Leftrightarrow P_{N_0|C_0,F_0,M_0}, & I_{C_0} \wedge I_{F_0} \wedge I_{M_0} \wedge I_{N_1} \Leftrightarrow P_{N_1|C_0,F_0,M_0}, \\ & & \ldots, & \ldots, \\ & & I_{C_1} \wedge I_{F_1} \wedge I_{M_1} \wedge I_{N_0} \Leftrightarrow P_{N_0|C_1,F_1,M_1}, & I_{C_1} \wedge I_{F_1} \wedge I_{M_1} \wedge I_{N_1} \Leftrightarrow P_{N_1|C_1,F_1,M_1}. \end{array}$$

- A weight is assigned to each literal in the propositional formula. Each positive literal of a parameter variable is assigned a weight equal to the corresponding probability entry in the CPT table,

$$\begin{array}{lll} C & : & weight(P_{C_0}) = P(C=0), \\ & & weight(P_{C_1}) = P(C=1), \\ F & : & weight(P_{F_0}) = P(F=0), \\ & & weight(P_{F_1}) = P(F=1), \\ M & : & weight(P_{M_0}) = P(M=0), \\ & & weight(P_{M_1}) = P(M=1), \\ N & : & weight(P_{N_0|C_0,F_0,M_0}) = P(N=0 \mid C=0, F=0, M=0), \\ & & weight(P_{N_1|C_0,F_0,M_0}) = P(N=1 \mid C=0, F=0, M=0), \\ & & \ldots, \\ & & weight(P_{N_0|C_1,F_1,M_1}) = P(N=0 \mid C=1, F=1, M=1), \\ & & weight(P_{N_1|C_1,F_1,M_1}) = P(N=1 \mid C=1, F=1, M=1). \end{array}$$

All other literals (both positive and negative) are assigned a weight of 1; i.e., $weight(I_{C_0}) = weight(\neg I_{C_0}) = \cdots = weight(I_{N_1}) = weight(\neg I_{N_1}) = 1$ and $weight(\neg P_{C_0}) = \cdots =$ $weight(\neg P_{N_1|C_1,F_1,M_1}) = 1$. The basic idea is that the indicator variables specify the state of the world—i.e., a value for each random variable in the Bayesian network—and then the weights of the literals multiplied together give the probability of that state of the world.

Sang, Beame, and Kautz (2005a) introduced an alternative encoding of a Bayesian network into weighted model counting of a CNF formula. Sang et al.'s encoding creates fewer variables and clauses, but the size of generated clauses of multi-valued variables can be larger. As with Chavira and Darwiche's encoding presented above, we illustrate Sang et al.'s encoding using the Bayesian network shown in Figure 2, once again assuming the random variables are all Boolean and omitting the node *Headache*.

- As in Chavira and Darwiche's encoding, for each node, *indicator variables* are created and *indicator clauses* are generated which ensure that in each model exactly one of the corresponding indicator variables for each node is true.
- Let the values of the nodes be linearly ordered. For each CPT entry $P(Y = y \mid \boldsymbol{X})$ such that $y$ is not the last value in the domain of $Y$, a parameter variable $P_{y|\boldsymbol{X}}$ is created; e.g.,

$$\begin{array}{lll} C : P_{C_0}, & N : P_{N_0|C_0,F_0,M_0}, & P_{N_0|C_0,F_0,M_1}, \\ F : P_{F_0}, & P_{N_0|C_0,F_1,M_0}, & P_{N_0|C_0,F_1,M_1}, \\ M : P_{M_0}, & P_{N_0|C_1,F_0,M_0}, & P_{N_0|C_1,F_0,M_1}, \\ & P_{N_0|C_1,F_1,M_0}, & P_{N_0|C_1,F_1,M_1}. \end{array}$$

- For each CPT entry $P(Y = y_i \mid \boldsymbol{X})$, a parameter clause is generated. Let the ordered domain of $Y$ be $\{y_1, \ldots, y_k\}$ and let $\boldsymbol{X} = x_1, \ldots, x_l$. If $y_i$ is not the last value in the domain of $Y$, the clause is given by,

$$I_{x_1} \wedge \cdots \wedge I_{x_l} \wedge \neg P_{y_1|\boldsymbol{X}} \wedge \cdots \wedge \neg P_{y_{i-1}|\boldsymbol{X}} \wedge P_{y_i|\boldsymbol{X}} \Rightarrow I_{y_i}.$$

If $y_i$ is the last value in the domain of $Y$, the clause is given by,

$$I_{x_1} \wedge \cdots \wedge I_{x_l} \wedge \neg P_{y_1|\boldsymbol{X}} \wedge \cdots \wedge \neg P_{y_{k-1}|\boldsymbol{X}} \Rightarrow I_{y_k}.$$

For our running example, the following parameter clauses would be generated,

$$\begin{array}{lll} C : & P_{C_0} \Rightarrow I_{C_0} & \neg P_{C_0} \Rightarrow I_{C_1} \\ F : & P_{F_0} \Rightarrow I_{F_0} & \neg P_{F_0} \Rightarrow I_{F_1} \\ M : & P_{M_0} \Rightarrow I_{M_0} & \neg P_{M_0} \Rightarrow I_{M_1} \\ N : & I_{C_0} \wedge I_{F_0} \wedge I_{M_0} \wedge P_{N_0|C_0,F_0,M_0} \Rightarrow I_{N_0}, & I_{C_0} \wedge I_{F_0} \wedge I_{M_0} \wedge \neg P_{N_0|C_0,F_0,M_0} \Rightarrow I_{N_1}, \\ & \ldots, & \ldots, \\ & I_{C_1} \wedge I_{F_1} \wedge I_{M_1} \wedge P_{N_0|C_1,F_1,M_1} \Rightarrow I_{N_0}, & I_{C_1} \wedge I_{F_1} \wedge I_{M_1} \wedge \neg P_{N_0|C_1,F_1,M_1} \Rightarrow I_{N_1}. \end{array}$$

- A weight is assigned to each literal in the propositional formula. As in Chavira and Darwiche's encoding, the weight of literals for indicator variables is always 1. The weight of literals for each parameter variable $P_{y|\boldsymbol{X}}$ is given by,

$$\begin{aligned} weight(P_{y|\boldsymbol{X}}) &= P(y \mid \boldsymbol{X}), \\ weight(\neg P_{y|\boldsymbol{X}}) &= 1 - P(y \mid \boldsymbol{X}). \end{aligned}$$

Let $F$ be the CNF encoding of a Bayesian network (either Chavira and Darwiche's encoding or Sang et al.'s encoding). A general query $P(\boldsymbol{Q} \mid \boldsymbol{E})$ on the network can be answered by,

$$\frac{weight(F \wedge Q \wedge E)}{weight(F \wedge E)}, \tag{7}$$

where $Q$ and $E$ are propositional formulas which enforce the appropriate values for the indicator variables that correspond to the known values of the random variables.

A backtracking algorithm used to enumerate the (weighted) models of a CNF formula is often referred to as DPLL or DPLL-based (in honor of Davis, Putnam, Logemann, and Loveland, the authors of some of the earliest work in the field: Davis & Putnam, 1960; Davis, Logemann, & Loveland, 1962), and usually includes such techniques as unit propagation, conflict recording, backjumping, and component caching.

## 3. Related Work

In this section, we relate our work to previously proposed methods for exact inference in Bayesian networks that contain noisy-OR/MAX relations.

One method for solving such networks is to replace each noisy-OR/MAX by its full CPT representation and then use any of the well-known algorithms for answering probabilistic queries such as variable elimination or tree clustering/jointree. However, in general—and in particular, for the networks that we use in our experimental evaluation—this method is impractical. A more fruitful approach for solving such networks is to take advantage of the structure or the semantics of the noisy-OR/MAX relations to improve both time and space efficiency (e.g., Heckerman, 1989; Olesen et al., 1989; D'Ambrosio, 1994; Heckerman & Breese, 1996; Zhang & Poole, 1996; Takikawa & D'Ambrosio, 1999; Díez & Galán, 2003; Chavira et al., 2005).

Quickscore (Heckerman, 1989) was the first efficient exact inference algorithm for Boolean-valued *two-layer* noisy-OR networks. Chavira, Allen and Darwiche (2005) present a method for *multi-layer* noisy-OR networks and show that their approach is significantly faster than Quickscore on randomly generated two-layer networks. Their approach proceeds as follows: (i) transform the noisy-OR network into a Bayesian network with full CPTs using Pearl's decomposition (see Figure 4), (ii) translate the network with full CPTs into CNF using a *general* encoding (see Section 2), (iii) simplify the resulting CNF by taking advantage of determinism (zero parameters and one parameters), and (iv) compile the CNF into an arithmetic circuit. One of our encodings for the noisy-OR (called WMC1) is similar to their more indirect (but also more general) proposal for encoding noisy-ORs (steps (i)–(iii)). We perform a detailed comparison in Section 4.1. In our experiments, we perform a detailed empirical comparison of their approach using compilation (steps (i)–(iv)) against Cachet using our encodings on large Bayesian networks.

Many alternative methods have been proposed to decompose a noisy-OR/MAX by adding hidden or auxiliary nodes and then solving using adaptations of variable elimination or tree clustering (e.g., Olesen et al., 1989; D'Ambrosio, 1994; Heckerman & Breese, 1996; Takikawa & D'Ambrosio, 1999; Díez & Galán, 2003).

Olesen et al. (1989) proposed to reduce the size of the distribution for the OR/MAX operator by decomposing a deterministic OR/MAX node with $n$ parents into a set of binary OR/MAX operators. The method, called parent divorcing, constructs a binary tree by adding auxiliary nodes $Z_i$ such that $Y$ and each of the auxiliary nodes has exactly two parents. Heckerman (1993) presented a sequential decomposition method again based on adding auxiliary nodes $Z_i$ and decomposing into binary MAX operators. Here one constructs a linear decomposition tree. Both methods require similar numbers of auxiliary nodes and similarly sized CPTs. However, as Takikawa and D'Ambrosio (1999) note, using either parent divorcing or sequential decomposition, many decomposition trees can be constructed from the same original network—depending on how the causes are ordered—and the efficiency of query answering can vary exponentially when using variable elimination or tree clustering, depending on the particular query and the choice of ordering.

To take advantage of causal independence models, Díez (1993) proposed an algorithm for the noisy-MAX/OR. By introducing one auxiliary variable $Y'$, Díez's method leads to a complexity of $O(nd^2)$ for singly connected networks, where $n$ is the number of causes and $d$ is the size of the domains of the random variables. However, for networks with loops it needs to be integrated with local conditioning. Takikawa and D'Ambrosio (1999) proposed a similar multiplicative factorization approach. The complexity of their approach is $O(\max(2^d, nd^2))$. However, Takikawa and D'Ambrosio's approach allows more efficient elimination orderings in the variable elimination algorithm, while Díez's method enforces more restrictions on the orderings. More recently, Díez and Galán (2003) proposed a multiplicative factorization that improves on this previous work, as it has the advantages of both methods. We use their auxiliary graph as the starting point for the remaining three of our CNF encodings (WMC2, MAX1, and MAX2). In our experiments, we perform a detailed empirical comparison of their approach using variable elimination against our proposals on large Bayesian networks.

In our work, we build upon the DPLL-based weighted model counting approach of Sang, Beame, and Kautz (2005a). Their general encoding assumes full CPTs and yields a parameter clause for each CPT parameter. However, this approach is impractical for large-scale noisy-OR/MAX networks. Our special-purpose encodings extend the weighted model counting approach for exact inference to networks that were previously intractable for the approach.

## 4. Efficient Encodings of Noisy-OR into CNF

In this section, we present techniques for improving the weighted model counting approach for Bayesian networks with noisy-OR relations. In particular, we present two CNF encodings for noisy-OR relations that exploit their structure or semantics. For the noisy-OR relation we take advantage of the Boolean domains to simplify the encodings. We use as a running example the Bayesian network shown in Figure 2. In the subsequent section, we generalize to the noisy-MAX relation.

### 4.1 Weighted CNF Encoding 1: An Additive Encoding

Let there be causes $X_1, \ldots, X_n$ leading to an effect $Y$ and let there be a noisy-OR relation at node $Y$ (see Figure 1), where all random variables are assumed to have Boolean-valued domains.

In our first weighted model encoding method (WMC1), we introduce an indicator variable $I_Y$ for $Y$ and an indicator variable $I_{X_i}$ for each parent of $Y$. We also introduce a parameter variable $P_{q_i}$ for each parameter $q_i$, $1 \leq i \leq n$ in the noisy-OR (see Equation 1). The weights of these variables are as follows,

$$\begin{aligned} weight(I_{X_i}) = weight(\neg I_{X_i}) &= 1, \\ weight(I_Y) = weight(\neg I_Y) &= 1, \\ weight(P_{q_i}) &= 1 - q_i, \\ weight(\neg P_{q_i}) &= q_i. \end{aligned}$$

The noisy-OR relation can then be encoded as the formula,

$$(I_{X_1} \wedge P_{q_1}) \vee (I_{X_2} \wedge P_{q_2}) \vee \cdots \vee (I_{X_n} \wedge P_{q_n}) \Leftrightarrow I_Y. \tag{8}$$

The formula can be seen to be an encoding of Pearl's well-known decomposition for noisy-OR (see Figure 4).

**Example 6.** *Consider once again the Bayesian network shown in Figure 2 and the parameters for the noisy-ORs shown in Table 1. The WMC1 encoding introduces the five Boolean indicator variables $I_C$, $I_F$, $I_M$, $I_N$, and $I_H$, each with weight 1; and the six parameter variables $P_{0.6}$, $P_{0.5}$, $P_{0.4}$, $P_{0.3}$, $P_{0.2}$, and $P_{0.1}$, each with $weight(P_{q_i}) = 1 - q_i$ and $weight(\neg P_{q_i}) = q_i$. Using Equation 8, the noisy-OR at node Nausea can be encoded as,*

$$(I_C \wedge P_{0.6}) \vee (I_F \wedge P_{0.5}) \vee (I_M \wedge P_{0.4}) \Leftrightarrow I_N.$$

*To illustrate the weighted model counting of the formula, suppose that nausea and malaria are absent and cold and flu are present (i.e., Nausea = 0, Malaria = 0, Cold = 1, and Flu = 1; and for the corresponding indicator variables $I_N$ and $I_M$ are false and $I_C$ and $I_F$ are true). The formula can be simplified to,*

$$(P_{0.6}) \vee (P_{0.5}) \Leftrightarrow 0.$$

*There is just one model for this formula, the model that sets $P_{0.6}$ to false and $P_{0.5}$ to false. Hence, the weighted model count of this formula is $weight(\neg P_{0.6}) \times weight(\neg P_{0.5}) = 0.6 \times 0.5 = 0.3$, which is just the entry in the penultimate row of the full CPT shown in Example 2.*

Towards converting Equation 8 into CNF, we also introduce an auxiliary indicator variable $w_i$ for each conjunction such that $w_i \Leftrightarrow I_{X_i} \wedge P_{q_i}$. This dramatically reduces the number of clauses generated. Equation 8 is then transformed into,

$$
\begin{aligned}
(\neg I_Y \quad \vee \quad & ((w_1 \vee \cdots \vee w_n) \wedge \\
& (\neg I_{X_1} \vee \neg P_{q_1} \vee w_1) \wedge \\
& (I_{X_1} \vee \neg w_1) \wedge \\
& (P_{q_1} \vee \neg w_1) \\
& \wedge \cdots \wedge \\
& (\neg I_{X_n} \vee \neg P_{q_n} \vee w_n) \wedge \\
& (I_{X_n} \vee \neg w_n) \wedge \\
& (P_{q_n} \vee \neg w_n))) \wedge \\
(I_Y \quad \vee \quad & ((\neg I_{X_1} \vee \neg P_{q_1}) \\
& \wedge \cdots \wedge \\
& (\neg I_{X_n} \vee \neg P_{q_n}))).
\end{aligned}
$$

The formula is not in CNF, but can be easily transformed into CNF using the distributive law. It can be seen that the WMC1 encoding can also easily encode evidence—i.e, if $I_Y = 0$ or $I_Y = 1$, the formula can be further simplified—before the final translation into CNF. Note that we have made the definitions of the auxiliary variables (i.e., $w_i \Leftrightarrow I_{X_i} \wedge P_{q_i}$) conditional on $I_Y$ being true, rather than just introducing separate clauses to define each auxiliary variable. This allows the formula to be further simplified in the presence of evidence and only introduces the $w_i$ if they are actually needed. In particular, if we know that $I_Y$ is false, all of the clauses involving the auxiliary variables $w_i$, including the definitions of the $w_i$, disappear when the formula is simplified.

**Example 7.** *Consider once again the Bayesian network shown in Figure 2. To illustrate the encoding of evidence, suppose that nausea is present (i.e., Nausea* $= 1$*) and headache is not present (i.e., Headache* $= 0$*). The corresponding constraints for the evidence are as follows.*

$$(I_C \wedge P_{0.6}) \vee (I_F \wedge P_{0.5}) \vee (I_M \wedge P_{0.4}) \Leftrightarrow 1 \tag{9}$$

$$(I_C \wedge P_{0.3}) \vee (I_F \wedge P_{0.2}) \vee (I_M \wedge P_{0.1}) \Leftrightarrow 0 \tag{10}$$

*The above constraints can be converted into CNF clauses. Constraint Equation 9 gives the clauses,*

$$
\begin{aligned}
& (w_1 \vee w_2 \vee w_3) \\
\wedge\ & (\neg I_C \vee \neg P_{0.6} \vee w_1) \wedge (I_C \vee \neg w_1) \wedge (P_{0.6} \vee \neg w_1) \\
\wedge\ & (\neg I_F \vee \neg P_{0.5} \vee w_2) \wedge (I_F \vee \neg w_2) \wedge (P_{0.5} \vee \neg w_2) \\
\wedge\ & (\neg I_M \vee \neg P_{0.4} \vee w_3) \wedge (I_M \vee \neg w_3) \wedge (P_{0.4} \vee \neg w_3)
\end{aligned}
$$

*and constraint Equation 10 gives the clauses,*

$$(\neg I_C \vee \neg P_{0.3}) \wedge (\neg I_F \vee \neg P_{0.2}) \wedge (\neg I_M \vee \neg P_{0.1}).$$

To show the correctness of encoding WMC1 of a noisy-OR, we first show that each entry in the full CPT representation of a noisy-OR relation can be determined using the weighted model count of the encoding. As always, let there be causes $X_1, \ldots, X_n$ leading to an effect $Y$ and let there be a noisy-OR relation at node $Y$, where all random variables have Boolean-valued domains.

**Lemma 1.** *Each entry in the full CPT representation of a noisy-OR at a node $Y$, $P(Y = y \mid X_1 = x_1, \ldots, X_n = x_n)$, can be determined using the weighted model count of Equation 8 created using the encoding WMC1.*

*Proof.* Let $F_Y$ be the encoding of the noisy-OR at node $Y$ using WMC1 and let $s$ be the set of assignments to the indicator variables $I_Y, I_{X_1}, \ldots, I_{X_n}$ corresponding to the desired entry in the CPT (e.g., if $Y = 0$, $I_Y$ is instantiated to false; otherwise it is instantiated to true). For each $X_i = 0$, the disjunct $(I_{X_i} \wedge P_{q_i})$ in Equation 8 is false and would be removed in the residual formula $F_Y|_s$; and for each $X_i = 1$, the disjunct reduces to $(P_{q_i})$. If $I_Y = 0$, each of the disjuncts in Equation 8 must be false and there is only a single model of the formula. Hence,

$$weight(F_Y|_s) = \prod_{i \in T_x} weight(\neg P_{q_i}) = \prod_{i \in T_x} q_i = P(Y = 0 \mid \boldsymbol{X}),$$

where $T_x = \{i \mid X_i = 1\}$ and $P(Y = 0 \mid \boldsymbol{X}) = 1$ if $T_x$ is empty. If $I_Y = 1$, at least one of the disjuncts in Equation 8 must be true and there are, therefore, $2^{|T_x|} - 1$ models. It can be seen that if we sum over all $2^{|T_x|}$ possible *assignments*, the weight of the formula is 1. Hence, subtracting off the one possible assignment that is not a model gives,

$$weight(F_Y|_s) = 1 - \prod_{i \in T_x} weight(\neg P_{q_i}) = 1 - \prod_{i \in T_x} q_i = P(Y = 1 \mid \boldsymbol{X}).$$

□

A *noisy-OR Bayesian network* over a set of random variables $Z_1, \ldots, Z_n$ is a Bayesian network where there are noisy-OR relations at one or more of the $Z_i$ and full CPTs at the remaining nodes. The next step in the proof of correctness is to show that each entry in the joint probability distribution represented by a noisy-OR Bayesian network can be determined using weighted model counting. In what follows, we assume that noisy-OR nodes are encoded using WMC1 and the remaining nodes are encoded using Sang et al.'s general encoding discussed in Section 2.2. Similar results can be stated using Chavira and Darwiche's general encoding.

**Lemma 2.** *Each entry in the joint probability distribution, $P(Z_1 = z_1, \ldots Z_n = z_n)$, represented by a noisy-OR Bayesian network can be determined using weighted model counting and encoding WMC1.*

*Proof.* Let $F$ be the encoding of the Bayesian network using WMC1 for the noisy-OR nodes and let $s$ be the set of assignments to the indicator variables $I_{Z_1}, \ldots, I_{Z_n}$ corresponding to the desired entry in the joint probability distribution. Any entry in the joint probability distribution can be expressed as the product,

$$P(X_1, \ldots, X_n) = \prod_{i=1}^{n} P(X_i \mid \mathit{parents}(X_i)),$$

where $n$ is the size of the Bayesian network and $\mathit{parents}(X_i)$ is the set of parents of $X_i$ in the directed graph; i.e., the entry in the joint probability distribution is determined by multiplying the corresponding CPT entries. For those nodes with full CPTs, $s$ determines the correct entry in each CPT by Lemma 2 in Sang et al. (2005a) and for those nodes with noisy-ORs, $s$ determines the correct probability by Lemma 1 above. Thus, $\mathit{weight}(F \wedge s)$ is the multiplication of the corresponding CPT entries; i.e., the entry in the joint probability distribution. □

The final step in the proof of correctness is to show that queries of interest can be correctly answered.

**Theorem 1.** *Given a noisy-OR Bayesian network, general queries of the form $P(\boldsymbol{Q} \mid \boldsymbol{E})$ can be determined using weighted model counting and encoding WMC1.*

*Proof.* Let $F$ be the CNF encoding of a noisy-OR Bayesian network. A general query $P(\boldsymbol{Q} \mid \boldsymbol{E})$ on the network can be answered by,

$$\frac{P(\boldsymbol{Q} \wedge \boldsymbol{E})}{P(\boldsymbol{E})} = \frac{\mathit{weight}(F \wedge Q \wedge E)}{\mathit{weight}(F \wedge E)},$$

where $Q$ and $E$ are propositional formulas that enforce the appropriate values for the indicator variables that correspond to the known values of the random variables. By definition, the function *weight* computes the weighted sum of the solutions of its argument. By Lemma 2, this is equal to the sum of the probabilities of those sets of assignments that satisfy the restrictions $Q \wedge E$ and $E$, respectively, which in turn is equal to the sum of the entries in the joint probability distribution that are consistent with $\boldsymbol{Q} \wedge \boldsymbol{E}$ and $\boldsymbol{E}$, respectively. □

As Sang et al. (2005a) note, the weighted model counting approach supports queries and evidence in arbitrary propositional form and such queries are not supported by any other exact inference method.

Our WMC1 encoding for noisy-OR is essentially similar to a more indirect but also more general proposal by Chavira and Darwiche (2005) (see Darwiche, 2009, pp. 313-323 for a detailed exposition of their proposal). Their approach proceeds as follows: (i) transform the noisy-OR network into a Bayesian network with full CPTs using Pearl's decomposition (see Figure 4), (ii) translate the network with full CPTs into CNF using a general encoding (see Section 2), and (iii) simplify the resulting CNF by taking advantage of determinism. Simplifying the resulting CNF proceeds as follows. Suppose we have in an encoding the sentence $(I_a \wedge I_{\neg b}) \Leftrightarrow P_{\neg b|a}$. If the parameter corresponding to $P_{\neg b|a}$ is zero, the sentence can be replaced by $\neg(I_a \wedge I_{\neg b})$ and $P_{\neg b|a}$ can be removed from the encoding. If the parameter corresponding to $P_{\neg b|a}$ is one, the entire sentence can be removed from the encoding. Applying their method to a noisy-OR (see Figure 1) results in the following,

$$
\begin{array}{lll}
 & & (\neg I_{X_1} \vee \neg P_{q_1} \vee w_1) \wedge \\
 & & (\neg I_{X_1} \vee P_{q_1} \vee \neg w_1) \wedge \\
 & & (I_{X_1} \vee \neg P_{q_1} \vee \neg w_1) \wedge \\
 & & (I_{X_1} \vee P_{q_1} \vee \neg w_1) \wedge \\
 & & \ldots \\
 & & (\neg I_{X_n} \vee \neg P_{q_n} \vee w_n) \wedge \\
 & & (\neg I_{X_n} \vee P_{q_n} \vee \neg w_n) \wedge \\
 & & (I_{X_n} \vee \neg P_{q_n} \vee \neg w_n) \wedge \\
 & & (I_{X_n} \vee P_{q_n} \vee \neg w_n) \wedge \\
(\neg I_Y & \vee & (w_1 \vee \cdots \vee w_n)) \wedge \\
(I_Y & \vee & ((\neg w_1 \vee \cdots \vee \neg w_{n-1} \vee \neg w_n) \wedge \\
 & & (\neg w_1 \vee \cdots \vee \neg w_{n-1} \vee w_n) \wedge \\
 & & \ldots \\
 & & (w_1 \vee \cdots \vee w_{n-1} \vee \neg w_n))),
\end{array}
$$

where we have simplified the expression by substituting equivalent literals and by using the fact that the random variables are Boolean (e.g., we use $I_{X_1}$ and $\neg I_{X_1}$ rather than $I_{X_1=0}$ and $I_{X_1=1}$). Three differences can be noted. First, in our encoding the definitions of the $w_i$ are conditional on $I_Y$ being true, rather than being introduced as separate clauses. Second, our definitions of the $w_i$ are more succinct. Third, in our encoding there are a linear number of clauses conditioned on $I_Y$ whereas in the Chavira et al. encoding there are $2^n - 1$ clauses. We note, however, that Chavira, Allen, and Darwiche (2005) discuss a direct translation of a noisy-OR to CNF based on Pearl's decomposition that is said to compactly represent the noisy-OR (i.e., not an exponential number of clauses), but the specific details of the CNF formula are not given.

### 4.2 Weighted CNF Encoding 2: A Multiplicative Encoding

Again, let there be causes $X_1, \ldots, X_n$ leading to an effect $Y$ and let there be a noisy-OR relation at node $Y$ (see Figure 1), where all random variables are assumed to have Boolean-valued domains.

Our second weighted model encoding method (WMC2) takes as its starting point Díez and Galán's (2003) directed auxiliary graph transformation of a Bayesian network with a noisy-OR/MAX relation. Díez and Galán note that for the noisy-OR relation, Equation (6) can be represented as a product of matrices,

$$
\begin{pmatrix} P(Y=0 \mid \boldsymbol{X}) \\ P(Y=1 \mid \boldsymbol{X}) \end{pmatrix} = \begin{pmatrix} 1 & 0 \\ -1 & 1 \end{pmatrix} \begin{pmatrix} P(Y \leq 0 \mid \boldsymbol{X}) \\ P(Y \leq 1 \mid \boldsymbol{X}) \end{pmatrix}.
$$

Based on this factorization, one can integrate a noisy-OR node into a regular Bayesian network by introducing a hidden node $Y'$ for each noisy-OR node $Y$. The transformation first creates a graph with the same set of nodes and arcs as the original network. Then, for each node $Y$ with a noisy-OR relation, we add a hidden node $Y'$ with the same domain as $Y$, add an arc $Y' \rightarrow Y$, redirect each arc $X_i \rightarrow Y$ to $X_i \rightarrow Y'$, and associate with $Y$ a factorization table,

| | $Y' = 0$ | $Y' = 1$ |
|---|---|---|
| $Y = 0$ | 1 | 0 |
| $Y = 1$ | $-1$ | 1. |

This auxiliary graph is not a Bayesian network as it contains parameters that are less than 0. So the CNF encoding methods for general Bayesian networks (see Section 2) cannot be applied here.

We introduce indicator variables $I_{Y'}$ and $I_Y$ for $Y'$ and $Y$, and an indicator variable $I_{X_i}$ for each parent of $Y'$. The weights of these variables are as follows,

$$\begin{aligned} weight(I_{Y'}) &= weight(\neg I_{Y'}) &= 1, \\ weight(I_Y) &= weight(\neg I_Y) &= 1, \\ weight(I_{X_i}) &= weight(\neg I_{X_i}) &= 1. \end{aligned}$$

For each arc $X_i \rightarrow Y'$, $1 \leq i \leq n$, we create two parameter variables $P^0_{X_i,Y'}$ and $P^1_{X_i,Y'}$. The weights of these variables are as follows,

$$\begin{aligned} weight(P^0_{X_i,Y'}) &= 1, & weight(P^1_{X_i,Y'}) &= q_i, \\ weight(\neg P^0_{X_i,Y'}) &= 0, & weight(\neg P^1_{X_i,Y'}) &= 1 - q_i. \end{aligned}$$

For each factorization table, we introduce two variables, $u_Y$ and $w_Y$, where the weights of these variables are given by,

$$\begin{aligned} weight(u_Y) &= 1, & weight(\neg u_Y) &= 0, \\ weight(w_Y) &= -1, & weight(\neg w_Y) &= 2. \end{aligned}$$

For the first row of a factorization table, we generate the clause,

$$(\neg I_{Y'} \vee I_Y), \tag{11}$$

and for the second row, we generate the clauses,

$$(\neg I_{Y'} \vee \neg I_Y \vee u_Y) \wedge (I_{Y'} \vee \neg I_Y \vee w_Y). \tag{12}$$

Finally, for every parent $X_i$ of $Y'$, we generate the clauses,

$$(I_{Y'} \vee I_{X_i} \vee P^0_{X_i,Y'}) \wedge (I_{Y'} \vee \neg I_{X_i} \vee P^1_{X_i,Y'}). \tag{13}$$

We now have a conjunction of clauses; i.e., CNF.

**Example 8.** *Consider once again the Bayesian network shown in Figure 2 and the parameters for the noisy-ORs shown in Table 1. The auxiliary graph transformation is shown in Figure 5. The WMC2 encoding introduces the seven Boolean indicator variables $I_C$, $I_F$, $I_M$, $I'_N$, $I_N$, $I'_H$, and $I_H$; the twelve parameter variables,*

$$\begin{array}{llll} P^0_{C,N'} & P^1_{C,N'} & P^0_{C,H'} & P^1_{C,H'} \\ P^0_{F,N'} & P^1_{F,N'} & P^0_{F,H'} & P^1_{F,H'} \\ P^0_{M,N'} & P^1_{M,N'} & P^0_{M,H'} & P^1_{M,H'}; \end{array}$$

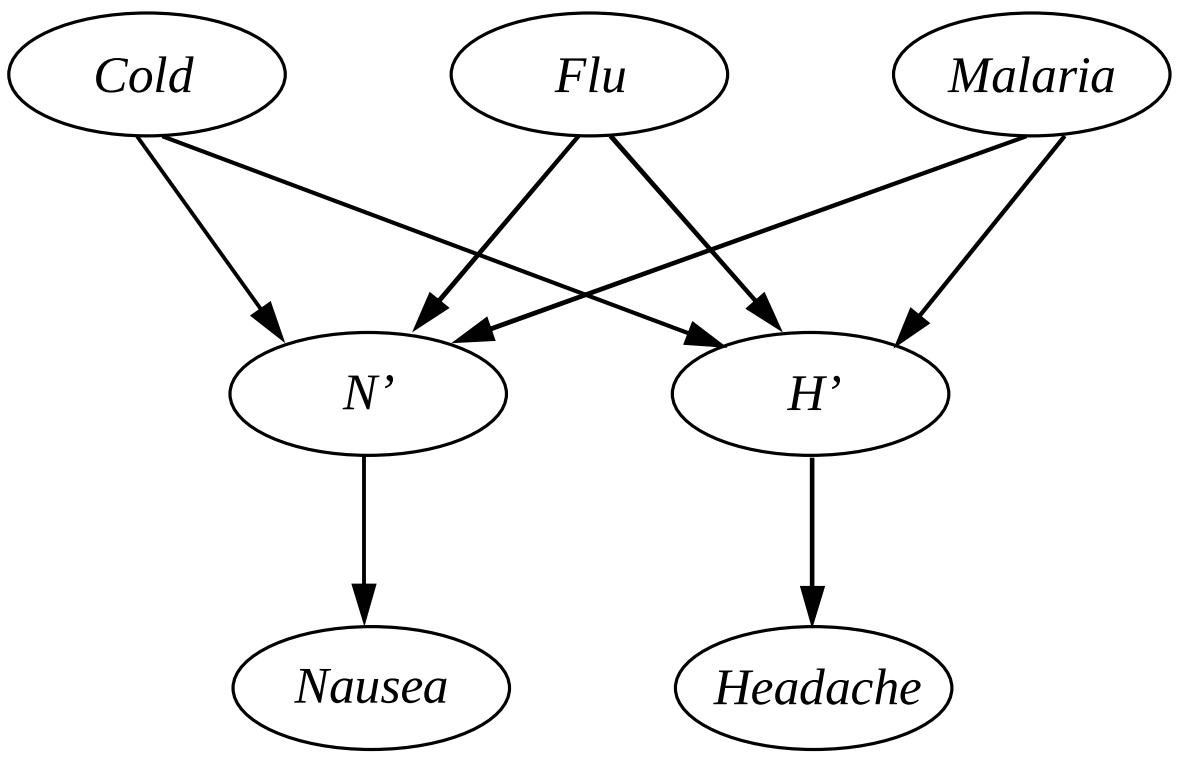

Figure 5: Díez and Galán's (2003) transformation of a noisy-OR relation applied to the Bayesian network shown in Figure 2.

*and the four factorization variables* $u_N$*,* $w_N$*,* $u_H$*, and* $w_H$*. The noisy-OR at node Nausea can be encoded as the set of clauses,*

$$\begin{array}{lll} \neg I_{N'} \vee I_N & I_{N'} \vee I_C \vee P^0_{C,N'} & I_{N'} \vee \neg I_C \vee P^1_{C,N'} \\ \neg I_{N'} \vee \neg I_N \vee u_N & I_{N'} \vee I_F \vee P^0_{F,N'} & I_{N'} \vee \neg I_F \vee P^1_{F,N'} \\ I_{N'} \vee \neg I_N \vee w_N & I_{N'} \vee I_M \vee P^0_{M,N'} & I_{N'} \vee \neg I_M \vee P^1_{M,N'} \end{array}$$

*To illustrate the weighted model counting of the formula, suppose that nausea and malaria are absent and cold and flu are present (i.e., Nausea* $= 0$*, Malaria* $= 0$*, Cold* $= 1$*, and Flu* $= 1$*; and for the corresponding indicator variables* $I_N$ *and* $I_M$ *are false and* $I_C$ *and* $I_F$ *are true). The formula can be simplified to,*

$$P^1_{C,N'} \wedge P^1_{F,N'} \wedge P^0_{M,N'}.$$

*(To see this, note that clauses that evaluate to true are removed and literals that evaluate to false are removed from a clause. As a result of simplifying the first clause,* $I_{N'}$ *is forced to be false and is removed from the other clauses.) There is just one model for this formula, the model that sets each of the conjuncts to true. Hence, the weighted model count of this formula is* $weight(P^1_{C,N'}) \times weight(P^1_{F,N'}) \times weight(P^0_{M,N'}) = 0.6 \times 0.5 \times 1.0 = 0.3$*, which is just the entry in the penultimate row of the full CPT shown in Example 2.*

Once again, it can be seen that WMC2 can also easily encode evidence into the CNF formula; i.e., if $I_Y = 0$ or $I_Y = 1$, the formula can be further simplified.

**Example 9.** *Consider once again the Bayesian network shown in Figure 2. To illustrate the encoding of evidence, suppose that nausea is present (i.e., Nausea* $= 1$*) and headache is not present (i.e., Headache* $= 0$*). The WMC2 encoding results in the following set of clauses,*

$$
\begin{array}{lll}
 & I_{N'} \vee I_C \vee P^0_{C,N'} & I_{N'} \vee \neg I_C \vee P^1_{C,N'} \\
\neg I_{N'} \vee u_N & I_{N'} \vee I_F \vee P^0_{F,N'} & I_{N'} \vee \neg I_F \vee P^1_{F,N'} \\
I_{N'} \vee w_N & I_{N'} \vee I_M \vee P^0_{M,N'} & I_{N'} \vee \neg I_M \vee P^1_{M,N'} \\
 & & \\
\neg I_{H'} & I_C \vee P^0_{C,H'} & \neg I_C \vee P^1_{C,H'} \\
 & I_F \vee P^0_{F,H'} & \neg I_F \vee P^1_{F,H'} \\
 & I_M \vee P^0_{M,H'} & \neg I_M \vee P^1_{M,H'}
\end{array}
$$

To show the correctness of encoding WMC2 of a noisy-OR, we first show that each entry in the full CPT representation of a noisy-OR relation can be determined using the weighted model count of the encoding. As always, let there be causes $X_1$, ..., $X_n$ leading to an effect $Y$ and let there be a noisy-OR relation at node $Y$, where all random variables have Boolean-valued domains.

**Lemma 3.** *Each entry in the full CPT representation of a noisy-OR at a node $Y$, $P(Y = y \mid X_1 = x_1, \ldots, X_n = x_n)$, can be determined using the weighted model count of Equations 11–13 created using the encoding WMC2.*

*Proof.* Let $F_Y$ be the encoding of the noisy-OR at node $Y$ using WMC2 and let $s$ be the set of assignments to the indicator variables $I_Y$, $I_{X_1}, \ldots, I_{X_n}$ corresponding to the desired entry in the CPT. For each $X_i = 0$, the clauses in Equation 13 reduce to $(I_{Y'} \vee P^0_{X_i,Y'})$, and for each $X_i = 1$, the clauses reduce to $(I_{Y'} \vee P^1_{X_i,Y'})$. If $I_Y = 0$, the clauses in Equations 11 & 12 reduce to $(\neg I_{Y'})$. Hence,

$$
\begin{aligned}
weight(F_Y|_s) &= weight(\neg I_{Y'}) \prod_{i \notin T_x} weight(P^0_{X_i,Y'})) \prod_{i \in T_x} weight(P^1_{X_i,Y'})) \\
&= \prod_{i \in T_x} q_i \\
&= P(Y = 0 \mid \boldsymbol{X}),
\end{aligned}
$$

where $T_x = \{i \mid X_i = 1\}$ and $P(Y = 0 \mid \boldsymbol{X}) = 1$ if $T_x$ is empty. If $I_Y = 1$, the clauses in Equations 11 & 12 reduce to $(\neg I_{Y'} \vee u_Y) \wedge (I_{Y'} \vee w_Y)$. Hence,

$$
\begin{aligned}
weight(F_Y|_s) &= weight(\neg I_{Y'}) weight(\neg u_Y) weight(w_Y) \prod_{i \in T_x} q_i + \\
&\quad weight(\neg I_{Y'}) weight(u_Y) weight(w_Y) \prod_{i \in T_x} q_i + \\
&\quad weight(I_{Y'}) weight(u_Y) weight(\neg w_Y) + \\
&\quad weight(I_{Y'}) weight(u_Y) weight(w_Y) \\
&= 1 - \prod_{i \in T_x} q_i \\
&= P(Y = 1 \mid \boldsymbol{X}).
\end{aligned}
$$

□

The remainder of the proof of correctness for encoding WMC2 is similar to that of encoding WMC1.

**Lemma 4.** *Each entry in the joint probability distribution, $P(Z_1 = z_1, \ldots Z_n = z_n)$, represented by a noisy-OR Bayesian network can be determined using weighted model counting and encoding WMC2.*

**Theorem 2.** *Given a noisy-OR Bayesian network, general queries of the form $P(\boldsymbol{Q} \mid \boldsymbol{E})$ can be determined using weighted model counting and encoding WMC2.*

## 5. Efficient Encodings of Noisy-MAX into CNF

In this section, we present techniques for improving the weighted model counting approach for Bayesian networks with noisy-MAX relations. In particular, we present two CNF encodings for noisy-MAX relations that exploit their structure or semantics. We again use as a running example the Bayesian network shown in Figure 2.

Let there be causes $X_1, \ldots, X_n$ leading to an effect $Y$ and let there be a noisy-MAX relation at node $Y$ (see Figure 1), where the random variables may have multi-valued (non-Boolean) domains. Let $d_X$ be the number of values in the domain of some random variable $X$.

The WMC2 multiplicative encoding above can be extended to noisy-MAX by introducing more indicator variables to represent variables with multiple values. In this section, we explain the extension and present two noisy-MAX encodings based on two different weight definitions of the parameter variables. The two noisy-MAX encodings are denoted MAX1 and MAX2, respectively. We begin by presenting those parts of the encodings that MAX1 and MAX2 have in common. As with WMC2, these two noisy-MAX encodings take as their starting point Díez and Galán's (2003) directed auxiliary graph transformation of a Bayesian network with noisy-OR/MAX. Díez and Galán show that for the noisy-MAX relation, Equation (6) can be factorized as a product of matrices,

$$P(Y = y \mid \boldsymbol{X}) = \sum_{y'=0}^{y} M_Y(y, y') \cdot P(Y \leq y' \mid \boldsymbol{X}) \tag{14}$$

where $M_Y$ is a $d_Y \times d_Y$ matrix given by,

$$M_Y(y, y') = \begin{cases} 1, & \text{if } y' = y, \\ -1, & \text{if } y' = y - 1, \\ 0, & \text{otherwise.} \end{cases}$$

For each noisy-MAX node $Y$, we introduce $d_Y$ indicator variables $I_{Y_0}$ ... $I_{Y_{d_Y-1}}$, to represent each value in the domain of $Y$, and $\binom{d_Y}{2} + 1$ clauses to ensure that exactly one of these variables is true. As in WMC2, we introduce a hidden node $Y'$ with the same domain as $Y$, corresponding indicator variables to represent each value in the domain of $Y'$, and clauses to ensure that exactly one domain value is selected in each model. For each parent $X_i$, $1 \leq i \leq n$, of $Y$, we define indicator variables $I_{i,x}$, where $x = 0, \ldots, d_{X_i} - 1$, and add clauses that ensure that exactly one of the indicator variables corresponding to each $X_i$ is true. Each indicator variable and each negation of an indicator variable has weight 1.

**Example 10.** *Consider once again the Bayesian network shown in Figure 2 and the parameters for the noisy-MAX shown in Table 2. As the node Nausea has domain {absent = 0, mild = 1, severe = 2} and the parents Cold, Flu, and Malaria are Boolean valued, both the MAX1 and MAX2 encodings introduce the Boolean indicator variables* $I_{N_a}$, $I_{N_m}$, $I_{N_s}$, $I_{N'_a}$, $I_{N'_m}$, $I_{N'_s}$, $I_{C_0}$, $I_{C_1}$, $I_{F_0}$, $I_{F_1}$, $I_{M_0}$, *and* $I_{M_1}$. *The weights of these variables and their negations are 1. Four clauses are added over the indicator variables for Nausea,*

$$\begin{aligned}(I_{N_a} \vee I_{N_m} \vee I_{N_s}) \quad & \wedge \quad (\neg I_{N_a} \vee \neg I_{N_m}) \\ & \wedge \quad (\neg I_{N_a} \vee \neg I_{N_s}) \\ & \wedge \quad (\neg I_{N_m} \vee \neg I_{N_s}).\end{aligned}$$

*Similar clauses are added over the indicator variables for the hidden node* $N'$ *and over the indicator variables for the parents Cold, Flu, and Malaria, respectively.*

For each factorization table, we introduce two auxiliary variables, $u_Y$ and $w_Y$, where the weights of these variables are given by,

$$\begin{aligned} weight(u_Y) &= 1, & weight(\neg u_Y) &= 0, \\ weight(w_Y) &= -1, & weight(\neg w_Y) &= 2. \end{aligned}$$

For each factorization table, a clause is added for each entry in the matrix,

$$M_Y(y, y') = \begin{cases} 1, & \text{add } (\neg I_{y'} \vee \neg I_y \vee u_Y) \quad \text{if } y' = y, \\ -1, & \text{add } (\neg I_{y'} \vee \neg I_y \vee w_Y) \quad \text{if } y' = y - 1, \\ 0, & \text{add } (\neg I_{y'} \vee \neg I_y \vee \neg u_Y) \quad \text{otherwise.} \end{cases}$$

**Example 11.** *Consider once again the Bayesian network shown in Figure 2 and the parameters for the noisy-MAX shown in Table 2. As Nausea has domain {absent = 0, mild = 1, severe = 2}, the factorization table* $M_N$ *is given by,*

| | $N' = absent$ | $N' = mild$ | $N' = severe$ |
|---|---|---|---|
| $N = absent$ | *1* | *0* | *0* |
| $N = mild$ | *−1* | *1* | *0* |
| $N = severe$ | *0* | *−1* | *1.* |

*Auxiliary variables* $u_N$ *and* $w_N$ *are introduced and the following clauses, shown in row order, would be added for the factorization table* $M_N$,

$$\begin{array}{lll} \neg I_{N_a} \vee \neg I_{N'_a} \vee u_N & \neg I_{N_a} \vee \neg I_{N'_m} \vee \neg u_N & \neg I_{N_a} \vee \neg I_{N'_s} \vee \neg u_N \\ \neg I_{N_m} \vee \neg I_{N'_a} \vee w_N & \neg I_{N_m} \vee \neg I_{N'_m} \vee u_N & \neg I_{N_m} \vee \neg I_{N'_s} \vee \neg u_N \\ \neg I_{N_s} \vee \neg I_{N'_a} \vee \neg u_N & \neg I_{N_s} \vee \neg I_{N'_m} \vee w_N & \neg I_{N_s} \vee \neg I_{N'_s} \vee u_N. \end{array}$$

That completes the description of those parts of the encodings that are common to both MAX1 and MAX2.

### 5.1 Weighted CNF Encoding 1 for Noisy-MAX

Our first weighted model counting encoding for noisy-MAX relations (MAX1) is based on an additive definition of noisy-MAX. Recall the decomposed probabilistic model for the noisy-MAX relation discussed at the end of Section 2.1. It can be shown that for the noisy-MAX, $P(Y \leq y \mid X_1, \ldots, X_n)$ can be determined using,

$$P(Y \leq y \mid X_1, \ldots, X_n) = \sum_{Y_i \leq y} \prod_{i=1}^{n} P(Y_i \mid X_i) = \sum_{Y_i \leq y} \prod_{\substack{i=1 \\ X_i \neq 0}}^{n} q_{i,Y_i}^{X_i} \tag{15}$$

where the $q_{i,Y_i}^{X_i}$ are the parameters to the noisy-MAX, and the sum is over all the configurations or possible values for $Y_1, \ldots, Y_n$, such that each of these values is less than or equal to the value $y$. Note that the outer operator is summation; hence, we refer to MAX1 as an additive encoding. Substituting the above into Equation 14 gives,

$$P(Y = y \mid X_1, \ldots, X_n) = \sum_{y'=0}^{y} M_Y(y, y') \cdot \left( \sum_{Y_i \leq y'} \prod_{\substack{i=1 \\ X_i \neq 0}}^{n} q_{i,Y_i}^{X_i} \right). \tag{16}$$

It is this equation that we encode into CNF. The encoding of the factorization table $M_Y$ is common to both encodings and has been explained above. It remains to encode the computation for $P(Y \leq y \mid X_1, \ldots, X_n)$.

For each parent $X_i$, $1 \leq i \leq n$, of $Y$ we introduce $d_Y$ indicator variables, $I_{i,y}$, to represent the effect of $X_i$ on $Y$, where $0 \leq y \leq d_Y - 1$, and add clauses that ensure that exactly one of the indicator variables correspond to each $X_i$ is true. Note that these indicators variables are in addition to the indicator variables common to both encodings and explained above. As always with indicator variables, the weights of $I_{i,y}$ and $\neg I_{i,y}$ are both 1.

For each parameter $q_{i,y}^{x}$ to the noisy-MAX, we introduce a corresponding parameter variable $P_{i,y}^{x}$. The weight of each parameter variable is given by,

$$weight(P_{i,y}^{x}) = q_{i,y}^{x} \qquad\qquad weight(\neg P_{i,y}^{x}) = 1$$

where $1 \leq i \leq n$, $0 \leq y \leq d_Y - 1$, and $1 \leq x \leq d_{X_i} - 1$. The relation between $X_i$ and $Y$ is represented by the parameter clauses[1],

$$(I_{i,x} \wedge I_{i,y}) \Leftrightarrow P_{i,y}^{x}$$

where $1 \leq i \leq n$, $0 \leq y \leq d_Y - 1$, and $1 \leq x \leq d_{X_i} - 1$.

**Example 12.** *Consider once again the Bayesian network shown in Figure 2 and the parameters for the noisy-MAX shown in Table 2. For the noisy-MAX at node Nausea, the encoding introduces the indicator variables* $I_{C,N_a}$, $I_{C,N_m}$, $I_{C,N_s}$, $I_{F,N_a}$, $I_{F,N_m}$, $I_{F,N_s}$, $I_{M,N_a}$, $I_{M,N_m}$, *and* $I_{M,N_s}$, *all with weight 1, and the clauses,*

1. To improve readability, in this section the propositional formulas are sometimes written in a more natural but non-clausal form. We continue to refer to them as clauses when the translation to clause form is straightforward.

$$\begin{array}{lll} I_{C,N_a} \vee I_{C,N_m} \vee I_{C,N_s} & I_{F,N_a} \vee I_{F,N_m} \vee I_{F,N_s} & I_{M,N_a} \vee I_{M,N_m} \vee I_{M,N_s} \\ \neg I_{C,N_a} \vee \neg I_{C,N_m} & \neg I_{F,N_a} \vee \neg I_{F,N_m} & \neg I_{M,N_a} \vee \neg I_{M,N_m} \\ \neg I_{C,N_a} \vee \neg I_{C,N_s} & \neg I_{F,N_a} \vee \neg I_{F,N_s} & \neg I_{M,N_a} \vee \neg I_{M,N_s} \\ \neg I_{C,N_m} \vee \neg I_{C,N_s} & \neg I_{F,N_m} \vee \neg I_{F,N_s} & \neg I_{M,N_m} \vee \neg I_{M,N_s} \end{array}$$

*As well, the following parameter variables and associated weights would be introduced,*

$$\begin{array}{lll} weight(P^1_{C,N_a}) = 0.7 & weight(P^1_{F,N_a}) = 0.5 & weight(P^1_{M,N_a}) = 0.1 \\ weight(P^1_{C,N_m}) = 0.2 & weight(P^1_{F,N_m}) = 0.2 & weight(P^1_{M,N_m}) = 0.4 \\ weight(P^1_{C,N_s}) = 0.1 & weight(P^1_{F,N_s}) = 0.3 & weight(P^1_{M,N_s}) = 0.5, \end{array}$$

*along with the following parameter clauses,*

$$\begin{array}{lll} (I_{C_1} \wedge I_{C,N_a}) \Leftrightarrow P^1_{C,N_a} & (I_{F_1} \wedge I_{F,N_a}) \Leftrightarrow P^1_{F,N_a} & (I_{M_1} \wedge I_{M,N_a}) \Leftrightarrow P^1_{M,N_a} \\ (I_{C_1} \wedge I_{C,N_m}) \Leftrightarrow P^1_{C,N_m} & (I_{F_1} \wedge I_{F,N_m}) \Leftrightarrow P^1_{F,N_m} & (I_{M_1} \wedge I_{M,N_m}) \Leftrightarrow P^1_{M,N_m} \\ (I_{C_1} \wedge I_{C,N_s}) \Leftrightarrow P^1_{C,N_s} & (I_{F_1} \wedge I_{F,N_s}) \Leftrightarrow P^1_{F,N_s} & (I_{M_1} \wedge I_{M,N_s}) \Leftrightarrow P^1_{M,N_s} \end{array}$$

It remains to relate (i) the indicator variables, $I_{i,x}$, which represent the value of the parent variable $X_i$, where $x = 0, \ldots, d_{X_i} - 1$; (ii) the indicator variables, $I_{i,y}$, which represent the effect of $X_i$ on $Y$, where $y = 0, \ldots, d_Y - 1$; and (iii) the indicator variables, $I_{Y'_{y'}}$, which represent the value of the hidden variable $Y'$, where $y' = 0, \ldots, d_Y - 1$. Causal independent clauses define the relation between (i) and (ii) and assert that if the cause $X_i$ is absent ($X_i = 0$), then $X_i$ has no effect on $Y$; i.e.,

$$I_{i,x_0} \Rightarrow I_{i,y_0}$$

where $1 \leq i \leq n$. Value constraint clauses define the relation between (ii) and (iii) and assert that if the hidden variable $Y'$ takes on a value $y'$, then the effect of $X_i$ on $Y$ cannot be that $Y$ takes on a higher degree or more severe value $y$; i.e.,

$$I_{Y'_{y'}} \Rightarrow \neg I_{i,Y_y}$$

where $1 \leq i \leq n$, $0 \leq y' \leq d_Y - 1$, and $y' < y \leq d_Y - 1$.

**Example 13.** *Consider once again the Bayesian network shown in Figure 2 and the parameters for the noisy-MAX shown in Table 2. For the noisy-MAX at node Nausea, the encoding introduces the causal independence clauses,*

$$I_{C_0} \Rightarrow I_{C,N_a} \qquad I_{F_0} \Rightarrow I_{F,N_a} \qquad I_{M_0} \Rightarrow I_{M,N_a}$$

*the value constraint clauses for $N' = absent$,*

$$\begin{array}{lll} I_{N'_a} \Rightarrow \neg I_{C,N_m} & I_{N'_a} \Rightarrow \neg I_{F,N_m} & I_{N'_a} \Rightarrow \neg I_{M,N_m} \\ I_{N'_a} \Rightarrow \neg I_{C,N_s} & I_{N'_a} \Rightarrow \neg I_{F,N_s} & I_{N'_a} \Rightarrow \neg I_{M,N_s} \end{array}$$

*and the value constraint clauses for $N' = mild$,*

$$I_{N'_m} \Rightarrow \neg I_{C,N_s} \qquad I_{N'_m} \Rightarrow \neg I_{F,N_s} \qquad I_{N'_m} \Rightarrow \neg I_{M,N_s}$$

### 5.2 Weighted CNF Encoding 2 for Noisy-MAX

Our second weighted model counting encoding for noisy-MAX relations (MAX2) is based on a multiplicative definition of noisy-MAX. Equation 5 states that $P(Y \leq y \mid X_1, \ldots, X_n)$ can be determined using,

$$P(Y \leq y \mid \boldsymbol{X}) = \prod_{\substack{i=1 \\ x_i \neq 0}}^{n} \sum_{y'=0}^{y} q_{i,y'}^{x_i}. \tag{17}$$

Note that the outer operator is multiplication; hence we refer to MAX2 as a multiplicative encoding. Substituting the above into Equation 14 gives,

$$P(Y = y \mid X_1, \ldots, X_n) = \sum_{y'=0}^{y} M_Y(y, y') \cdot \left( \prod_{\substack{i=1 \\ x_i \neq 0}}^{n} \sum_{y''=0}^{y'} q_{i,y''}^{x_i} \right). \tag{18}$$

It is this equation that we encode into CNF. The encoding of the factorization table $M_Y$ is common to both encodings and has been explained above. It remains to encode the computation for $P(Y \leq y \mid X_1, \ldots, X_n)$.

For each parameter $q_{i,y}^{x}$ to the noisy-MAX, we introduce a corresponding parameter variable $P_{i,y}^{x}$. The weight of each parameter variable pre-computes the summation in Equation 17,

$$weight(P_{i,y}^{x}) = \sum_{y'=0}^{y} q_{i,y'}^{x} \qquad\qquad weight(\neg P_{i,y}^{x}) = 1$$

where $1 \leq i \leq n$, $0 \leq y \leq d_Y - 1$, and $1 \leq x \leq d_{X_i} - 1$. The relation between $X_i$ and $Y'$ is represented by the parameter clauses,

$$(I_{i,x} \wedge I_{y'}) \Leftrightarrow P_{i,y}^{x},$$

where $0 \leq y \leq d_Y - 1$ and $0 \leq x \leq d_{X_i} - 1$.

**Example 14.** *Consider once again the Bayesian network shown in Figure 2 and the parameters for the noisy-MAX shown in Table 2. For the noisy-MAX at node Nausea, the following parameter variables and associated weights would be introduced,*

$$\begin{array}{lll} weight(P_{C,N_a}^{1}) = 0.7 & weight(P_{F,N_a}^{1}) = 0.5 & weight(P_{M,N_a}^{1}) = 0.1 \\ weight(P_{C,N_m}^{1}) = 0.9 & weight(P_{F,N_m}^{1}) = 0.7 & weight(P_{M,N_m}^{1}) = 0.5 \\ weight(P_{C,N_s}^{1}) = 1 & weight(P_{F,N_s}^{1}) = 1 & weight(P_{M,N_s}^{1}) = 1, \end{array}$$

*along with the following parameter clauses,*

$$\begin{array}{lll} (I_{C_1} \wedge I_{N'_a}) \Leftrightarrow P_{C,N_a}^{1} & (I_{F_1} \wedge I_{N'_a}) \Leftrightarrow P_{F,N_a}^{1} & (I_{M_1} \wedge I_{N'_a}) \Leftrightarrow P_{M,N_a}^{1} \\ (I_{C_1} \wedge I_{N'_m}) \Leftrightarrow P_{C,N_m}^{1} & (I_{F_1} \wedge I_{N'_m}) \Leftrightarrow P_{F,N_m}^{1} & (I_{M_1} \wedge I_{N'_m}) \Leftrightarrow P_{M,N_m}^{1} \\ (I_{C_1} \wedge I_{N'_s}) \Leftrightarrow P_{C,N_s}^{1} & (I_{F_1} \wedge I_{N'_s}) \Leftrightarrow P_{F,N_s}^{1} & (I_{M_1} \wedge I_{N'_s}) \Leftrightarrow P_{M,N_s}^{1} \end{array}$$

As stated so far, the encoding is sufficient for correctly determining each entry in the full CPT representation of a noisy-MAX relation using weighted model counting. However, to improve the efficiency of the encoding, we add redundant clauses. The redundant clauses do not change the set of solutions to the encoding, and thus do not change the weighted model count. They do, however, increase propagation and thus the overall speed of computation in the special case where all of the causes are absent. To this end, for each noisy-MAX node $Y$, we introduce an auxiliary variable $I_{v_Y}$ with weights given by,

$$weight(I_{v_Y}) \quad = \quad 1, \qquad weight(\neg I_{v_Y}) \quad = \quad 0,$$

and we introduce the clauses,

$$\left(\bigwedge_i^n I_{i,0}\right) \Rightarrow (I_{Y'_0} \Rightarrow I_{v_Y}), \qquad \left(\bigwedge_i^n I_{i,0}\right) \Rightarrow (I_{Y_0} \Rightarrow I_{v_Y}),$$

and the clauses,

$$\left(\bigwedge_i^n I_{i,0}\right) \Rightarrow (I_{y'} \Rightarrow \neg I_{v_Y}), \qquad \left(\bigwedge_i^n I_{i,0}\right) \Rightarrow (I_{y} \Rightarrow \neg I_{v_Y}),$$

where $1 \leq y' \leq d_Y - 1$ and $1 \leq y \leq d_Y - 1$.

**Example 15.** *Consider once again the Bayesian network shown in Figure 2. For the noisy-MAX at node Nausea, an auxiliary variable $I_{v_N}$ is introduced with $weight(I_{v_N}) = 1$ and $weight(\neg I_{v_N}) = 0$ along with the following redundant clauses,*

$$\begin{array}{ll}
(I_{C_0} \wedge I_{F_0} \wedge I_{M_0}) \Rightarrow (I_{N'_a} \Rightarrow I_{v_N}) & (I_{C_0} \wedge I_{F_0} \wedge I_{M_0}) \Rightarrow (I_{N_a} \Rightarrow I_{v_N}) \\
(I_{C_0} \wedge I_{F_0} \wedge I_{M_0}) \Rightarrow (I_{N'_m} \Rightarrow \neg I_{v_N}) & (I_{C_0} \wedge I_{F_0} \wedge I_{M_0}) \Rightarrow (I_{N_m} \Rightarrow \neg I_{v_N}) \\
(I_{C_0} \wedge I_{F_0} \wedge I_{M_0}) \Rightarrow (I_{N'_s} \Rightarrow \neg I_{v_N}) & (I_{C_0} \wedge I_{F_0} \wedge I_{M_0}) \Rightarrow (I_{N_s} \Rightarrow \neg I_{v_N}).
\end{array}$$

## 6. Experimental Evaluation

In this section, we empirically evaluate the effectiveness of our encodings. We use the Cachet solver[2] in our experiments as it is one of the fastest weighted model counting solvers. We compare against Ace (version 2) (Chavira et al., 2005) and Díez and Galán's (2003) approach using variable elimination.

We chose to compare against Ace for two reasons. First, Ace did well in the 2008 exact inference competition (no winner was declared, but Ace performed better on more classes of problems than all other entries). Second, other methods that are publicly available or that did well at the competition, such as Smile/GeNIe (Druzdzel, 2005) or Cachet using a general encoding on the full CPT representation, currently do not take any computational advantage of noisy-OR and noisy-MAX and thus would be "straw" algorithms. A strength of Ace is that it does take advantage of local structure and determinism and it specifically takes advantage of the semantics of the noisy-OR and noisy-MAX to speed up computation. The comparison to Ace, while revealing, is not without its methodological difficulties however (see Section 6.4 for a discussion).

2. `http://www.cs.rochester.edu/u/kautz/Cachet/index.htm`

We chose to compare against Díez and Galán's (2003) approach, which consists of variable elimination applied to an auxiliary network that permits exploitation of causal independence, as they show that the approach is more efficient than previous proposals for noisy-MAX. To our knowledge, this work has not been subsequently superseded; i.e., it is still the state-of-the-art on improving variable elimination for noisy-MAX for exact inference. No implementation of Díez and Galán's approach is publicly available, and so we implemented it ourselves. Our implementation uses algebraic decision diagrams (ADDs) (Bahar, Frohm, Gaona, Hachtel, Macii, Pardo, & Somenzi, 1993) as the base data structure to represent conditional probability tables. Algebraic decision diagrams permit a compact representation by aggregating identical probability values and speed up computation by exploiting context-specific independence (Boutilier, Friedman, Goldszmidt, & Koller, 1996), taking advantage of determinism and caching intermediate results to avoid duplicate computation. While ADDs are more complicated than table based representations, their ability to exploit structure often yields a speed up that is greater than the incurred overhead. In fact, ADDs are currently the preferred data structure for inference in factored partially observable Markov decision processes (Shani, Brafman, Shimony, & Poupart, 2008). The variable elimination heuristic that we used is a greedy one that first eliminates all variables that appear in deterministic potentials of one variable (this is equivalent to unit propagation) and then eliminates the variable that creates the smallest algebraic decision diagram with respect to the eliminated algebraic decision diagrams. In order to avoid creating an algebraic decision diagram for each variable when searching for the next variable to eliminate, the size of a new algebraic decision diagram is estimated by the smallest of two upper bounds: (i) the cross product of the domain size of the variables of the new algebraic decision diagram and (ii) the product of the sizes (e.g., the number of nodes) of the eliminated algebraic decision diagrams.

Good variable ordering heuristics play an important role in the success of modern DPLL-based model counting solvers. Here, we evaluate two heuristics: *Variable State Aware Decaying Sum* (VSADS) and *Tree Decomposition Variable Group Ordering* (DTree). The VSADS heuristic is one of the current best performing dynamic heuristics designed for DPLL-based model counting engines (Sang, Beame, & Kautz, 2005b). It can be viewed as a scoring system that attempts to satisfy the most recent conflict clauses and also considers the number of occurrences of a variable at the same time. Compared with the VSADS heuristic, the DTree heuristic (Huang & Darwiche, 2003) can be described as a mixed variable ordering heuristic. DTree first uses a binary tree decomposition to generate ordered variable groups. The decomposition is done prior to search. The order of the variables within a group is then decided dynamically during the backtracking search using a dynamic heuristic.

All of the experiments were performed on a Pentium workstation with 3GHz hyper-threading CPU and 2GB RAM.

### 6.1 Experiment 1: Random Two-Layer Networks

In our first set of experiments, we used randomly generated two-layer networks to compare the time and space efficiency of the WMC1 and WMC2 encodings.

Both the WMC1 and WMC2 encodings can answer probabilistic queries using Equation 7. Both encodings lead to quick factorization given evidence during the encoding. The clauses from negative evidence can be represented compactly in the resulting CNF, even with a large number of parents. In the WMC2 encoding, positive evidence can be represented by just three Boolean variables (see Example 9 for an illustration of which variables are deleted and which are kept for the case of positive evidence), whereas the WMC1 encoding requires $n$ Boolean variables, one for each parent (see Example 7). In the WMC2 encoding, we use two parameter variables ($P^0_{X_i,Y'}$ and $P^1_{X_i,Y'}$) to represent every arc, while the WMC1 encoding only needs one.

Table 3: Binary, two layer, noisy-OR networks with 500 diseases and 500 symptoms. Effect of increasing amount of positive evidence ($P^+$) on number of variables in encoding ($n$), treewidth of the encoding (width), average time to solve (sec.), and number of instances solved within a cutoff of one hour (solv.), where the test set contained a total of 30 instances. The results for $P^+ = 5, \ldots, 25$ are similar to the results for $P^+ = 30$ and are omitted.

| | WMC1 | | | | WMC2 | | | | Ace | |
|---|---|---|---|---|---|---|---|---|---|---|
| $P^+$ | $n$ | width | sec. | solv. | $n$ | width | sec. | solv. | sec. | solv. |
| 30 | 3686 | 10 | 0.2 | 30 | 6590 | 11 | 0.1 | 30 | 31.7 | 30 |
| 35 | 3716 | 11 | 0.6 | 30 | 6605 | 11 | 0.2 | 30 | 32.5 | 30 |
| 40 | 3746 | 13 | 21.4 | 30 | 6620 | 11 | 0.5 | 30 | 32.7 | 30 |
| 45 | 3776 | 14 | 38.8 | 30 | 6635 | 13 | 2.0 | 30 | 35.7 | 30 |
| 50 | 3806 | 19 | 75.3 | 30 | 6650 | 13 | 6.1 | 30 | 40.9 | 30 |
| 55 | 3836 | 22 | 175.2 | 30 | 6665 | 16 | 71.0 | 30 | 166.0 | 30 |
| 60 | 3916 | 24 | | 17 | 6680 | 16 | | 27 | | 21 |

Each random network contains 500 diseases and 500 symptoms. Each symptom has six possible diseases uniformly distributed in the disease set. Table 3 shows the treewidth of the encoded CNF for the WMC1 and WMC2 encodings. The first column shows the amount of positive evidence in the symptom variables. The remainder of the evidence variables are negative symptoms. It can be seen that although the WMC1 encoding generates fewer variables than the WMC2 encoding, the CNF created by the WMC2 encoding has smaller width. The probability of evidence (PE) is computed using the tree decomposition guided variable ordering (Huang & Darwiche, 2003) and the results are compared against Ace[3] (a more detailed experimental analysis is given in the next experiments).

### 6.2 Experiment 2: QMR-DT

In our second set of experiments, we used a Bayesian network called QMR-DT. In comparison to randomly generated problems, QMR-DT presents a real-world inference task with various structural and sparsity properties. For example, in the empirical distribution of diseases, a small proportion of the symptoms are connected with a large number of diseases (see Figure 6).

3. `http://reasoning.cs.ucla.edu/ace/`

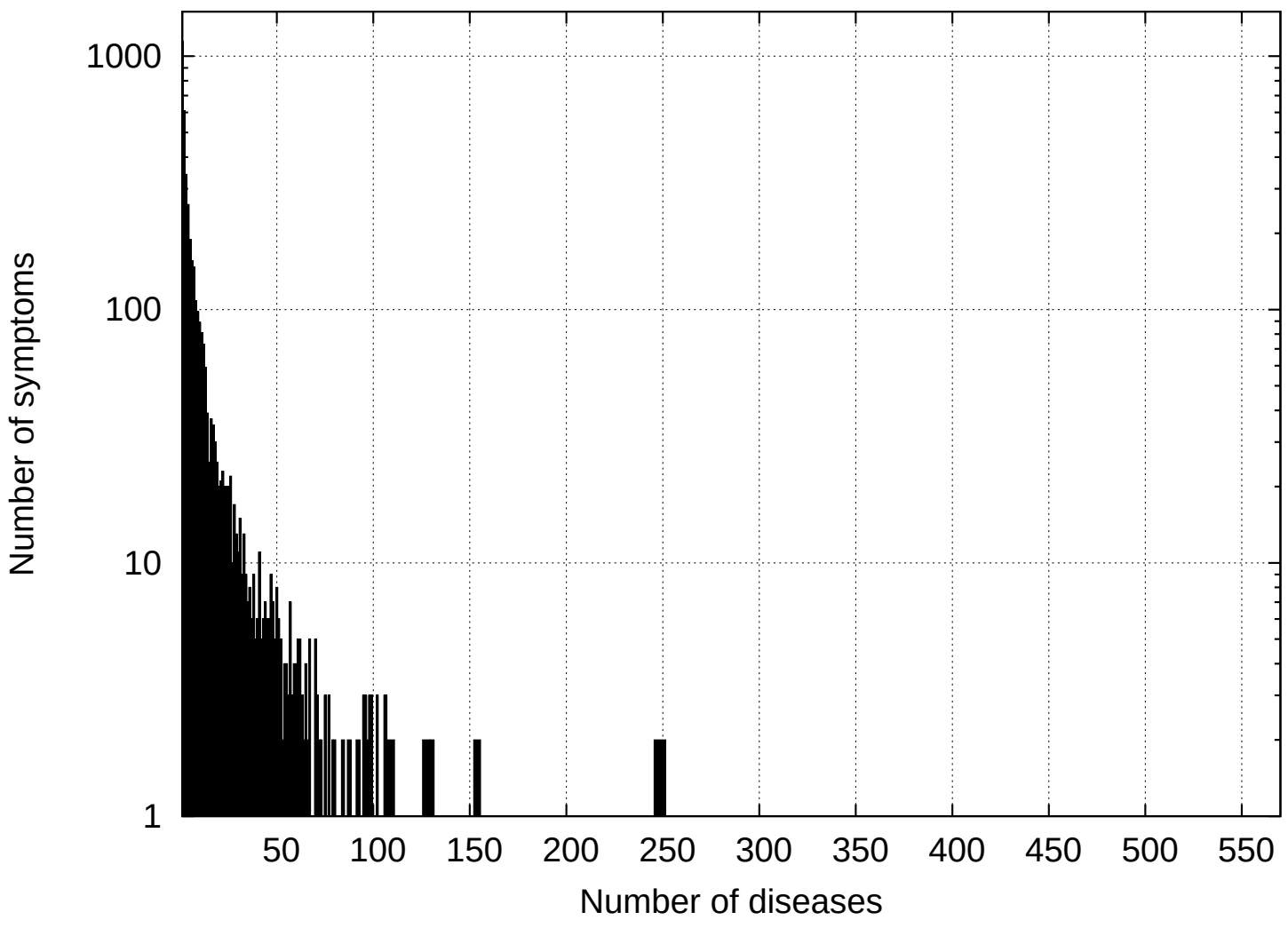

Figure 6: Empirical distribution of diseases in the QMR-DT Bayesian network. Approximately 80% of the symptoms are connected with less than 50 diseases.

The network we used was aQMR-DT, an anonymized version of QMR-DT[4]. Symptom vectors with $k$ positive symptoms were generated for each experiment. For each evidence vector, the symptom variables were sorted into ascending order by their parent (disease) number, the first $k$ variables were chosen as positive symptoms, and the remaining symptom variables were set to negative. The goal of the method is to generate instances of increasing difficulty.

We report the runtime to answer the probability of evidence (PE) queries. We also experimented with an implementation of Quickscore[5], but found that it could not solve any of the test cases shown in Figure 7. The approach based on weighted model counting also outperforms variable elimination on QMR-DT. The model counting time for 2560 positive symptoms, when using the WMC1 encoding and the VSADS dynamic variable ordering heuristic, is 25 seconds. This same instance could not be solved within one hour by variable elimination.

We tested two different heuristics on each encoding: the VSADS dynamic variable order heuristic and DTree (Huang & Darwiche, 2003), the semi-static tree decomposition-based heuristic. The runtime using an encoding and the DTree heuristic is the sum of two parts: the preprocessing time by DTree and the runtime of model counting on the encoding. In this experiment, DTree had a faster runtime than VSADS in the model counting process. However, the overhead of preprocessing for large size networks is too high to achieve better overall performance.

The WMC2 encoding generates twice as many variables as the WMC1 encoding. Although the WMC2 encoding is more promising than the WMC1 encoding on smaller size

4. `http://www.utoronto.ca/morrislab/aQMR.html`
5. `http://www.cs.ubc.ca/~murphyk/Software/BNT/bnt.html`

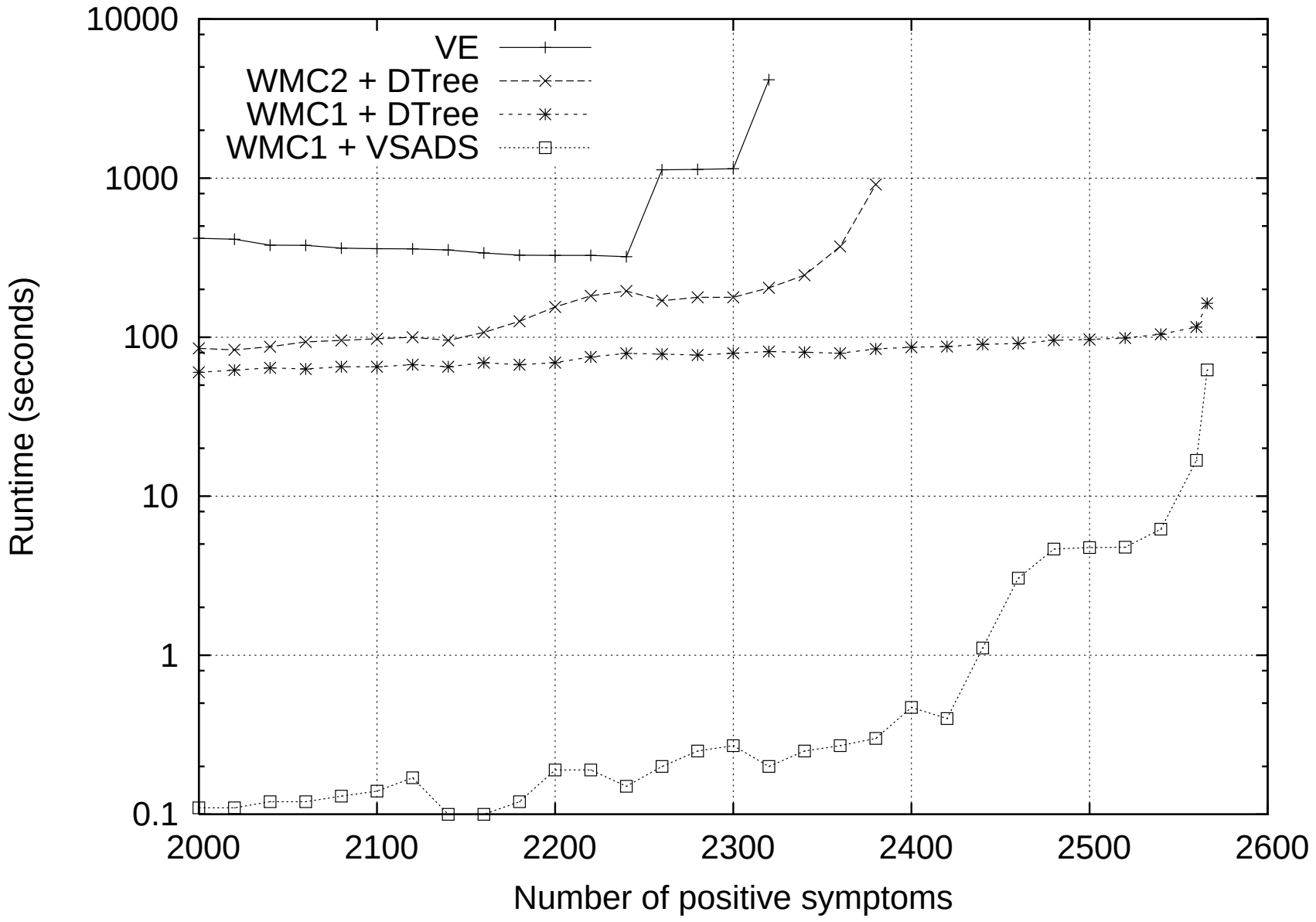

Figure 7: The QMR-DT Bayesian network with 4075 symptoms and 570 diseases. Effect of amount of positive symptoms on the time to answer probability of evidence queries, for the WMC1 encoding and the DTree variable ordering heuristic, the WMC1 encoding and the VSADS variable ordering heuristic, the WMC2 encoding and the DTree variable ordering heuristic, and Díez and Galán's (2003) approach using variable elimination. ACE could not solve instances with more than 500 positive symptoms within a one hour limit on runtime.

networks (see Table 3), here the WMC2 encoding is less efficient than the WMC1 encoding. The overhead of the tree decomposition ordering on the WMC2 encoding is also higher than on the WMC1 encoding. Our results also show that dynamic variable ordering does not work well on the WMC2 encoding. Model counting using the WMC2 encoding and the VSADS heuristic cannot solve networks when the amount of positive evidence is greater than 1500 symptoms.

The experimental results also show that our approach is more efficient than ACE. For example, using ACE, a CNF of QMR-DT with 30 positive symptoms creates $2.8 \times 10^5$ variables, $2.8 \times 10^5$ clauses and $3.8 \times 10^5$ literals. Also, it often requires more than 1GB of memory to finish the compilation process. With the WMC1 encoding, the same network and the same evidence create only $4.6 \times 10^4$ variables, $4.6 \times 10^4$ clauses and $1.1 \times 10^5$ literals. Cachet, the weighted model counting engine, needs less than 250MB of memory in most cases to solve these instances. And in our experiments, ACE could not solve QMR-DT with more than 500 positive symptoms within an hour.

### 6.3 Experiment 3: Random Multi-Layer Networks

In our third set of experiments, we used randomly generated multi-layer Bayesian networks. To test randomly generated multi-layer networks, we constructed a set of acyclic Bayesian networks using the same method as Díez and Galán (2003): create $n$ binary variables; randomly select $m$ pairs of nodes and add arcs between them, where an arc is added from $X_i$ to $X_j$ if $i < j$; and assign a noisy-OR distribution or a noisy-MAX distribution to each node with parents.

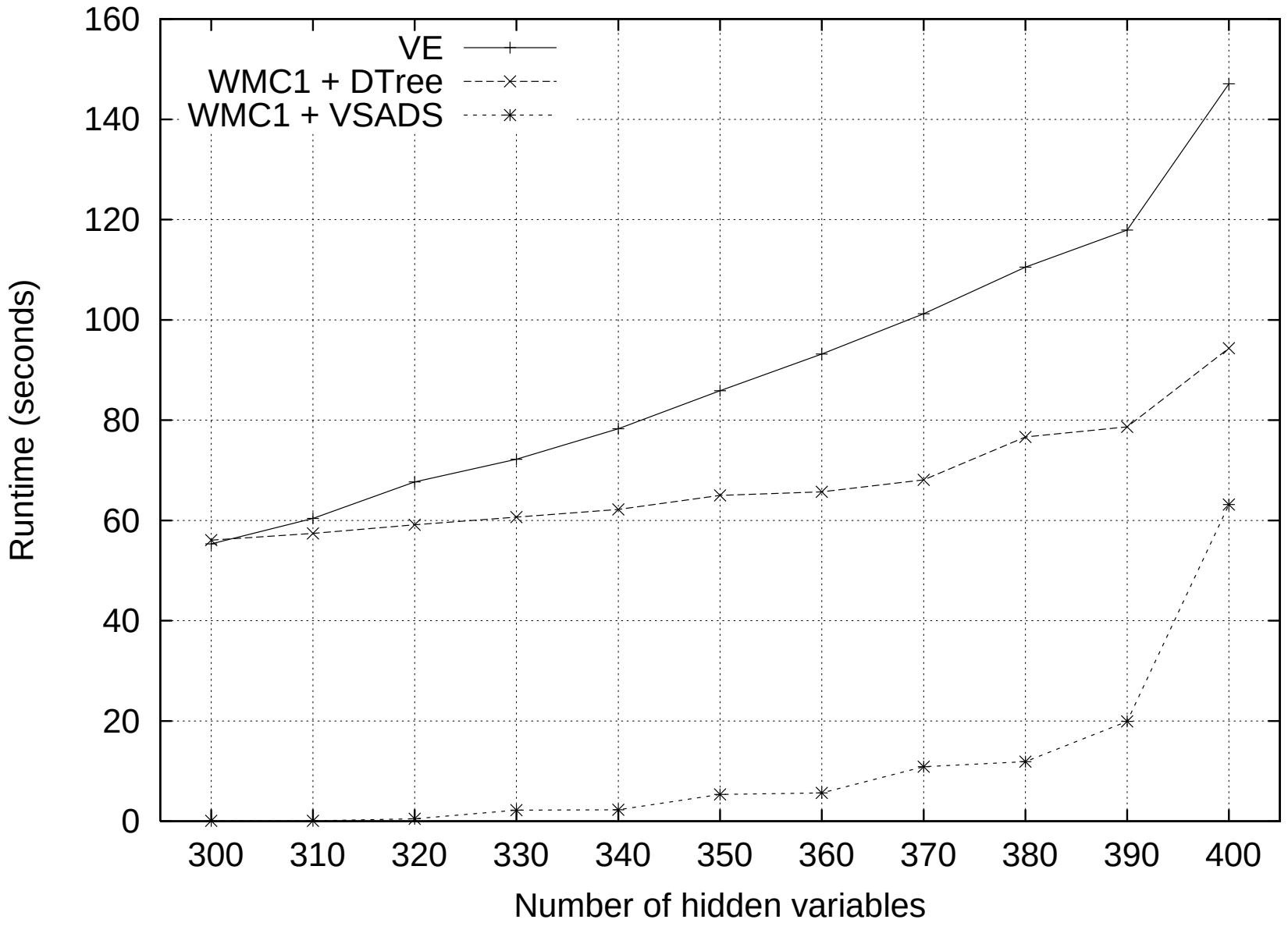

Figure 8: Random noisy-OR Bayesian networks with 3000 random variables. Effect of number of hidden variables on average time to answer probability of evidence queries, for the WMC1 encoding and the VSADS variable ordering heuristic, the WMC1 encoding and the DTree variable ordering heuristic, and Díez and Galán's (2003) approach using variable elimination.

Figure 8 shows the effect of the number of hidden variables on the average time to answer probability of evidence (PE) queries for random noisy-OR Bayesian networks. Each data point is an average over 30 randomly generated instances, where each instance had 3000 nodes in total.

The results from the two layer QMR-DT and the multiple layer random noisy-OR show that on average, the approach based on weighted model counting performed significantly better than variable elimination and significantly better than Ace. All the approaches benefit from the large amount of evidence, but the weighted model counting approach explores the determinism more efficiently with dynamic decomposition and unit propagation. In comparison to variable elimination, the weighted model counting approach encodes the local dependencies among parameters and the evidence into clauses/constraints. The topological or structural features of the CNF, such as connectivity, can then be explored dynamically during DPLL's simplification process.

Heuristics based primarily on conflict analysis have been successfully applied in modern SAT solvers. However, Sang, Beame, and Kautz (2005b) note that for model counting it is often the case that there are few conflicts in those parts of the search tree where there are large numbers of solutions and in these parts a heuristic based purely on conflict analysis will make nearly random decisions. Sang et al.'s (2005b) VSADS heuristic, which combines both conflict analysis and literal counting, avoids this pitfall and can be seen to work very well on these large Bayesian networks with large amounts of evidence. DTree is also a good choice due to its divide-and-conquer nature. However, when we use DTree to decompose the CNF generated from QMR-DT, usually the first variable group contains more than 500 disease variables. As well, the overhead of preprocessing affects the overall efficiency of this approach.

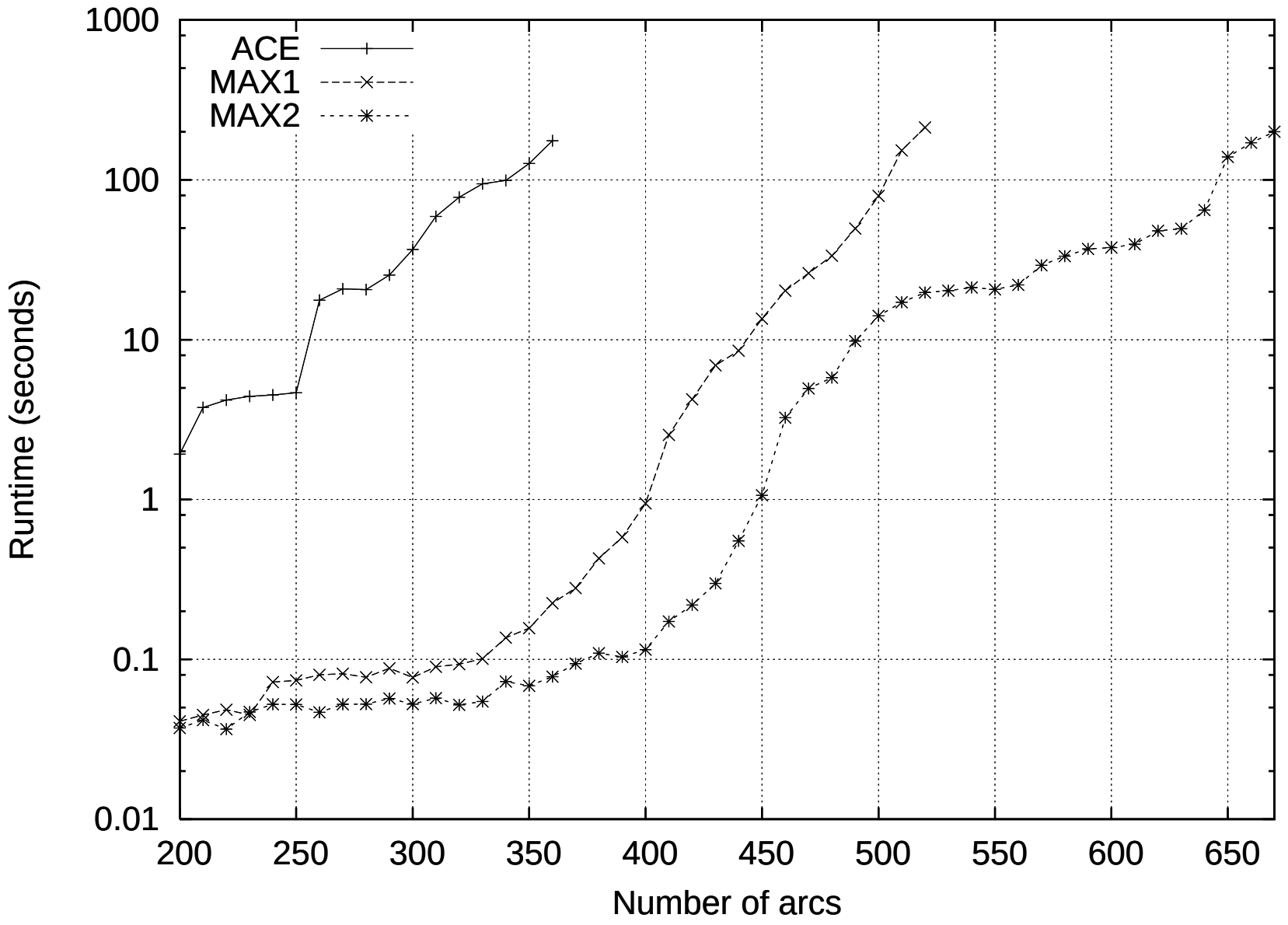

Figure 9: Random noisy-MAX Bayesian networks with 100 five-valued random variables. Effect of number of arcs on average time to answer probability of evidence queries, for the MAX1 encoding for noisy-MAX, the MAX2 encoding for noisy-MAX, and Chavira, Allen, and Darwiche's Ace (2005).

Similarly, we performed an experiment with 100 five-valued random variables. Figure 9 shows the effect of the number of arcs on the average time to answer probability of evidence (PE) queries for random noisy-MAX Bayesian networks. Each data point is an average over 50 randomly generated instances. It can be seen that on these instances our CNF encoding MAX2 out performs our encoding MAX1 and significantly outperforms Chavira, Allen, and Darwiche's Ace (2005). It has been recognized that for noisy-MAX relations, the multiplicative factorization has significant advantages over the additive factorization (Takikawa & D'Ambrosio, 1999; Díez & Galán, 2003). Hence, one would expect that the CNF encoding based on the multiplicative factorization (encoding MAX2) would perform better than the CNF encoding based on the additive factorization (encoding MAX1). The primary disadvantage of encoding MAX1 is that it must encode in the CNF summing over all configurations. As a result, MAX1 generates much larger CNFs than MAX2, including more variables and more clauses. In encoding MAX2, the weight of a parameter variable represents the maximum effect of each cause and hence minimizes the add computations.

### 6.4 Discussion

We experimentally evaluated our four SAT encodings—WMC1 and WMC2 for noisy-OR and MAX1 and MAX2 for noisy-MAX—on a variety of Bayesian networks using the Cachet weighted modeling counting solver. The WMC1 and MAX1 encodings can be characterized as additive encodings and the WMC2 and MAX2 encodings as multiplicative encodings. In our experiments, the multiplicative encodings gave SAT instances with smaller treewidth. For noisy-OR, the additive encoding (WMC1) gave smaller SAT instances than the multiplicative encoding (WMC2). For noisy-MAX, it was the reverse and the additive encoding (MAX1) gave larger SAT instances than the multiplicative encoding (MAX2). With regards to speedups, in the experiments for the noisy-OR, the results were mixed as to which encoding is better; sometimes it was WMC1 and other times WMC2. In the experiments for the noisy-MAX, the results suggest that the multiplicative encoding (MAX2) is better. Here the reduced treewidth and the reduced size of the MAX2 encoding were important, and WMC2 was able to solve many more instances.

We also compared against Díez and Galán's (2003) approach using variable elimination (hereafter, D&G) and against Ace (Chavira et al., 2005). In our experiments, our approach dominated D&G and Ace with speedups of up to three orders of magnitude. As well, our approach could solve many instances which D&G and Ace could not solve within the resource limits. However, our results should be interpreted with some care for at least three reasons. First, it is well known that the efficiency of variable elimination is sensitive to the variable elimination heuristic that is used and to how it is implemented. While we were careful to optimize our implementation and to use a high-quality heuristic, there is still the possibility that a different implementation or a different heuristic would lead to different results. Second, Cachet, which is based on search, is designed to answer a single query and our experiments are based on answering a single query. However, Ace uses a compilation strategy which is designed to answer multiple queries efficiently. The compilation step can take a considerable number of resources (both time and space) which does not payoff in our experimental design. Third, although Ace can be viewed as a weighted model counting solver, we are not comparing just encodings in our experiments. As Chavira and Darwiche (2008) note, Cachet and Ace differ in many ways including using different methods for decomposition, variable splitting, and caching. As well, Ace uses other optimizations that Cachet does not, including encoding equal parameters, eclauses (a succinct way of encoding that only one literal is true in a disjunction), and structured resolution. (We refer the reader to Chavira & Darwiche, 2008 for an experimental comparison of search and compilation, and an extensive discussion of the difficulty of comparing the two approaches and their advantages and disadvantages.) Nevertheless, in our experiments we demonstrated instances of noisy-OR networks (see Figure 7) and noisy-MAX networks (see Figure 9) that could not be solved at all by D&G and by Ace within the resource limits, but could be solved quite quickly by Cachet using our encodings.

## 7. Conclusions and Future Work

Large graphical models, such as QMR-DT, are often intractable for exact inference when there is a large amount of positive evidence. We presented time and space efficient CNF encodings for noisy-OR/MAX relations. We also explored alternative search ordering heuristics for the DPLL-based backtracking algorithm on these encodings. In our experiments, we showed that together our techniques extend the model counting approach for exact inference to networks that were previously intractable for the approach. As well, while our experimental results must be interpreted with some care as we are comparing not only our encodings but also implementations of systems with conflicting design goals, on these benchmarks our techniques gave speedups of up to three orders of magnitude over the best previous approaches and scaled up to larger instances. Future work could include developing specific CNF encodings of other causal independence relations (see Koller & Friedman, 2009, pp. 175–185).

## Acknowledgments

A preliminary version of this paper appeared as: Wei Li, Pascal Poupart, and Peter van Beek. Exploiting Causal Independence Using Weighted Model Counting. In *Proceedings of the 23rd AAAI Conference on Artificial Intelligence*, pages 337–343, 2008. The authors wish to thank the anonymous referees for their helpful comments.